\documentclass{article}

\usepackage{microtype}
\usepackage{graphicx}
\usepackage{subcaption}
\usepackage{booktabs} 
\usepackage{array}

\usepackage{hyperref}
\usepackage{comment}
\excludecomment{comment}

\usepackage[preprint]{icml2026}

\usepackage{amsmath}
\usepackage{amssymb}
\usepackage{mathtools}
\usepackage{relsize} 
\usepackage{amsthm}
\usepackage{enumitem}
\setlist[itemize]{itemsep=5pt, topsep=0pt, parsep=0pt, partopsep=0pt, leftmargin=*}
\setlist[enumerate]{itemsep=5pt, topsep=0pt, parsep=0pt, partopsep=0pt, leftmargin=*}
\usepackage{multirow}
\usepackage{subcaption}
\usepackage{tikz}
\usetikzlibrary{arrows}
\usetikzlibrary{shapes}
\usepackage{mystyle}
\usepackage[table]{xcolor}

    \usepackage{booktabs}
\usepackage{colortbl}
\usepackage{caption}
\DeclareCaptionFont{small}{\small}
\usepackage[capitalize,noabbrev]{cleveref}

\theoremstyle{plain}
\newtheorem{theorem}{Theorem}[section]
\newtheorem{proposition}[theorem]{Proposition}
\newtheorem{lemma}[theorem]{Lemma}
\newtheorem{corollary}[theorem]{Corollary}
\newtheorem{definition}[theorem]{Definition}

\theoremstyle{remark}
\newtheorem{remark}[theorem]{Remark}

\icmltitlerunning{Node-level generalization from a single graph}

\begin{document}

\twocolumn[
  \icmltitle{A theory of inductive node-level generalization from a single large graph}



  \icmlsetsymbol{equal}{*}

  \begin{icmlauthorlist}
    \icmlauthor{Nicolas Keriven}{nico}
    \icmlauthor{Yann Viegas}{yann}
  \end{icmlauthorlist}

  \icmlaffiliation{nico}{CNRS, IRISA, Rennes, France.}
  \icmlaffiliation{yann}{Ecole Normale Supérieure}

  \icmlcorrespondingauthor{Nicolas Keriven}{nicolas.keriven@cnrs.fr}

  \icmlkeywords{Machine Learning, ICML}

  \vskip 0.3in
]



\printAffiliationsAndNotice{}  

\begin{abstract}
  In Graph Machine Learning (GML), the notion of generalization fundamentally varies depending on the task level and the properties of the dataset. For instance, graph-level tasks fit into the traditional ML framework of i.i.d. samples, with relatively small graphs. On the contrary, node-level tasks involve dependent nodes, often in a few large graphs. Existing studies often consider the transductive settings, where training and test data co-exist in the same graph, or, for inductive learning, try to recover some kind of probabilistic ``independence'', e.g. by still assuming a large number of independent graphs or subgraphs. 

  In this work, we develop a novel framework to express the generalization capabilities of Graph ML algorithms for inductive node-level tasks when learning from a single large graph, taking inspiration from the literature on GNN transferability. In particular, we emphasize the similarities with classical ML by revisiting the excess risk decomposition with a new notion of ``transferable'' risk on large graphs. We apply this framework to Latent Position random graphs: we uncover new insights on the hardness of GML compared to traditional ML, and derive novel notions of concentration. Finally, we apply this framework to several models of GNNs, unifying several existing results on GNNs transferability and universality.
\end{abstract}

\section{Introduction}

Graph Machine Learning \cite{Bronstein2021,Hu2020, Morris2024a} (GML) is now a well-established field, with numerous applications 
 and a vast literature on its most popular class of models, Graph Neural Networks (GNN).
Naturally, the foremost question of traditional Machine Learning (ML), that of \emph{generalization} to unseen new data, has been extensively explored both theoretically and empirically \cite{Verma2019b, Garg2020, Vasileiou2025}. GML however does come with its own difficulties when attempting to apply classical ML theory. In particular, for \textbf{node-level prediction}, most of the usual tools break down: first, the main ML assumption of observing i.i.d. data is no longer valid, which destroys tools such as Rademacher symmetrization. Classically, this has been often handled by assuming a transductive learning context \cite{Esser2021, Tang2023}, or i.i.d. sampling of ego-subgraphs as GNN's input \cite{Verma2019b, generalization_gnn_layer}. Secondly, and often overlooked in existing literature, the notion of \emph{functions} is somewhat loose for node tasks. Many studies will attempt to manipulate ``functions'' from node features to labels, but often in an unrigorous manner. For instance, \cite{Lv2021} define a ``function space'' on nodes, computes its Rademacher complexity, then 
link it to generalization, however, as far as we can tell, their proof is incorrect, 
since the so-called ``symmetrization lemma'' \citep[Chap 4]{Bach2021} does not apply anymore. 

Due to these difficulties, generalization analyses in graph ML have been mostly limited to a) graph-level tasks \cite{Garg2020}, which fit in the traditional framework of i.i.d. samples, b) \emph{transductive} node tasks \cite{Esser2021}, where labelled and unlabelled nodes co-exist in the same graphs, or c) inductive node-level but with a large number of graphs to induce sufficient independence \cite{morris_node_generalization}. 
In contrast, in this paper, we introduce a theoretical framework to study \textbf{inductive node-level generalization} where the training data consist of a \textbf{single large graph}. Unsurprisingly, we will be heavily inspired by the notion of \emph{transferability} \cite{Levie2021, Ruiz2020b} between graph sizes, here generalized for our purpose. 

\paragraph{Outline.} We first formally integrate transferability into the classical ML framework in Sec.~\ref{sec:gml_risk} and define the class of \textbf{risk-transferable} equivariant functions over graphs, which is arguably the weakest form of transferability necessary in a node-task ML context. We formulate a new risk decomposition around this notion. 
Second, in Sec.~\ref{sec:lpm} we instantiate this framework for a general family of latent-position random graphs that includes SBMs, random geometric graphs, graphons, and so on. This allows us to draw new links between GML and traditional ML: we show that GML on the nodes can be exactly linked to \emph{regular ML on the latent variables}, but that it is \emph{provably harder}. Finally, in Sec.~\ref{sec:gnn} we instantiate the framework on several models of GNNs, mostly by drawing on existing transferability and universality results. 
All technical proofs are in Appendix. 

\paragraph{Related work.} The literature on GNN generalization is naturally extensive. As mentioned above, GNN generalization has been mostly studied for \emph{graph tasks}, which fit in regular ML with i.i.d. samples. 
As such, the usual arsenal of ML can be deployed, adapted to the structure of the space of graphs: covering numbers \cite{Levie2023, Rauchwerger2024, Maskey2022, Maskey2024}, Rademacher complexities \cite{Garg2020, Lv2021}, VC-dimension \cite{ScarselliVC, Morris2023, Franks2024, gene_covered_forest}, PAC-Bayes approaches \cite{Liao2021, Ju2023}... See \cite{Vasileiou2025} for a survey.

On the contrary, nodes (or link) tasks do not necessarily involve i.i.d. samples, and require the definition of a proper framework. For instance, many works adopt the \emph{transductive} settings \cite{Esser2021, Tang2023, gnngeneralization_spectral}, in which labelled and unlabelled nodes are in the same graph. The randomness then often comes from the labelled node sampling, and specialized tools such as ``transductive Rademacher'' have been developed \cite{transductive_rademacher, Esser2021}. 
For inductive node-tasks, there is arguably no consensus on a single theoretical framework. Some works assume that the $k$-hop subgraphs necessary for GNNs are somewhat drawn i.i.d. \cite{Verma2019b, generalization_gnn_layer, Li_generalization_homo}, which is a strong assumption. Closer to us, \cite{morris_node_generalization} consider non-i.i.d. node samples, where the graph structure encodes the statistical dependence. However their generalization bound degrades with the number of nodes that come from the same graph and is moot when training on a single graph. The authors in \cite{Lv2021} present a Rademacher-based approach, however, to the best of our knowledge, with a slightly incorrect way of using the symmetrization Lemma \citep[Chap 4]{Bach2021}. The work closest to this paper is \cite{Wang_manifold_generalization}, where the authors uses the transferability of Manifold NN to study generalization between graphs sampled from the same manifold. The present paper extends this approach to a general framework, and draws new connections with traditional ML generalization bounds, including universality. 

Our approach is strongly inspired by works on GNN \emph{transferability} \cite{Levie2021, Ruiz2020b, Keriven2020a, transfer_transferability,transfer_gmn}, which loosely refers to the fact that GNNs can transfer their results across graph sizes. It can be formulated through convergence results to limit objects \cite{Levie2021, Ruiz2020b, Keriven2020a, convergence_ign, gnn_graphops} or through regularity of GNNs with respect to certain norms \cite{Maskey2021, Levie2023,transfer_transferability,transfer_gmn}. The notion of ``risk-transferability'' that we introduce in Sec.~\ref{sec:gml_risk} is somewhat the ``weakest'' possible, as it only expresses transferability of the expected risk, as in \cite{transfer_gmn}. We relate it to convergence results in Sec.~\ref{sec:lpm}, and leave other transferability approaches for future work. 





\section{A statistical framework for inductive generalization from a single large graph}\label{sec:gml_risk}

\paragraph{Settings.} A graph $G=(A,Z)$ of size $n$ is defined by its weighted adjacency matrix $A \in \RR^{n \times n}$ (often binary) and node features matrix $Z \in \RR^{n \times d_z}$. We consider a framework where the training data is a \emph{single} graph $G$ with every node labelled (for simplicity), i.e. we observe $Y \in \Yy^n$. The goal is to characterize inductive prediction on new graphs. 
We start with a quick reminder on statistical machine learning.

\subsection{Background on Statistical Learning}\label{sec:course_ml}


In ``classical'' supervised ML (see e.g. \cite{Bach2021}) the goal is to predict a value $y \in \Yy$ from an observed quantity $x \in \Xx$. 
For simplicity we will always assume that $\Xx$ is a compact in $\RR^d$. 
The vast majority of ML models can be modelled as \textbf{prediction functions} $f:\Xx \to \oYy$, where for convenience the predictions sometimes live in a different set $\oYy$ instead of directly $\Yy$ (e.g. logits for classification). We denote by $\Ff_{all} = \oYy^\Xx$ the set of all functions. 

\paragraph{Expected risk.} In ML, it is generally assumed that pairs $(x,y)$ are drawn from some joint distribution $P_{xy}$.
Given some \emph{loss function} $\ell:\oYy \times \Yy \to \RR_+$, the quality of a prediction function $f$ is measured with the \textbf{expected risk}
\begin{equation}
    \label{eq:expectedrisk}
    \RrML(f) = \EE_{(x,y) \sim P_{xy}} \ell(f(x),y)\, .
\end{equation}
Then, the ultimate goal of ML is to minimize 
\begin{equation}
    \label{eq:fstar}
     \RrML^\star = \inf_{f \in \Ff_{all}} \RrML(f), \quad f^\star \in \uargmin{f \in \Ff_{all}} \RrML(f)\, .
\end{equation}
The quantity $\RrML^\star$ is called the Bayes risk, and when it exists, $f^\star$ is a Bayes predictor.

We will assume that the loss is either Lipschitz for some metric $d_\oYy$ over $\oYy$, e.g. the Categorical Cross Entropy (CCE) is Lipschitz when $\oYy$ is the simplex, or that it is the square loss (which is not technically Lipschitz), e.g. for regression.
\begin{assump}
    \label{ass:loss}
    The loss function $\ell$ is either:
    \begin{enumerate}
    \item $L_\ell$-Lipschitz in its first argument 
    \begin{equation*}
        \label{eq:assloss}
        \forall \hat y, \hat y' \in \oYy, y \in \Yy,~~\abs{\ell(\hat y, y) - \ell(\hat y', y)}\leq L_\ell d_\oYy(\hat y, \hat y') 
    \end{equation*}
    \item the square loss and $\Yy\subset \RR^{d_y}$ is bounded by $\norm{y} \leq D_\Yy$. In this case $\oYy = \RR^{d_y}$ and $d_{\oYy}$ is the Euclidean norm.
    \end{enumerate}
\end{assump}

\paragraph{Empirical risk, function set.} In practice, the full distribution $P_{xy}$, is of course inaccessible. Instead, one observes a collection of labelled examples $(x_1,y_1), \ldots, (x_n,y_n)$ which is assumed to be \textbf{identically and independently distributed} (i.i.d.) according to $P_{xy}$. Thus, the vast majority of ML methods naturally follow \emph{Empirical Risk Minimization} (ERM), in which $\RrML$ is replaced by the \textbf{empirical risk} $\hRrML = \frac{1}{n}\sum_{i=1}^n \ell(f(x_i), y_i)$. Moreover, one generally cannot optimize over all functions, and therefore the minimization is carried over a subset $\Ff \subset \Ff_{all}$, which yields: 
\begin{equation}
    \label{eq:empiricalrisk}
    \hat f \in \uargmin{f \in \Ff}  \hRrML(f)
\end{equation}

\paragraph{Risk decomposition.} The full \emph{generalization error} is measured through the \emph{excess risk} 
$
    \EeML(\hat f) = \RrML(\hat f) - \RrML^\star \geq 0
$ 
It is generally decomposed in two terms: denoting $\RrML^\Ff = \inf_{f \in \Ff} \RrML(f)$, an easy derivation yields
\begin{equation*}\label{eq:excessriskdecompositionML}
    \EeML(\hat f) \leq 2\underbrace{\sup_{f \in \Ff} \abs{\hRrML(f) - \RrML(f)}}_{\text{estimation error}} + \underbrace{\RrML^\Ff - \RrML^\star}_{\text{approximation error}}
\end{equation*}
Here, the \emph{estimation error} materializes the fact that the true risk has been replaced by the empirical risk. 
While pointwise approximation $\hRrML(f) \approx \RrML(f)$ is obvious as a sum of iid variables, the difficulty resides in handling the \emph{uniform} bound over $\Ff$. Several classical approaches have been developped for this: covering numbers, Rademacher complexities, VC dimension... One notices that the bound increases with the size of $\Ff$, which often translates to a bound on some \emph{capacity} of the model class (e.g., bounded widths and weights in neural nets).

The second term is the \emph{approximation error}, which evaluates how much functions in $\Ff$ can approximate the ideal function $f^\star$. It is highly problem-dependent. For instance, neural networks are known to be \textbf{universal} among continuous functions \cite{Cybenko1989}, such that the approximation error vanishes if $f^\star$ is continuous \emph{and the capacity of the model class increases}. One thus obtains a trade-off to balance the two error terms, even though obtaining explicit rates for the approximation error is generally difficult.

\subsection{A new risk for inductive node-tasks}\label{sec:newrisk}

We now revisit these notions for inductive node-tasks on graphs. Here, a prediction function $h(A,Z)$ takes a graph $G=(A,Z)$ and returns $Y\in \oYy^n$, that is, a label $y_i \in \oYy$ per node. 
For inductive node-tasks, where future input graphs are unknown, these functions must satisfy two conditions: a) they accept input $G$ \emph{of any size $n$}, and map them to an output $Y$ \emph{of the same size} $n$, and b) they are \emph{permutation-equivariant}: $h(\sigma A \sigma^\top, \sigma Z) = \sigma h(A,Z)$ for any permutation matrix $\sigma$.
We will call $\Hheq$ all such functions. GNNs are naturally in $\Hheq$: a GNN can be applied to graphs of any size, and they are permutation-equivariant by construction.

\paragraph{Empirical risk.} Unlike classical ML, the notion of ``expected'' risk is not immediate here. It is easier to start by defining the empirical risk that is \emph{actually} minimized in practice, and work from there. Recalling that the training dataset is comprised of \emph{one} graph $G = (A,Z)$ of size $n$ with all its nodes labelled, the empirical risk is 
\begin{equation}\label{eq:empiricalriskGML}
    \mathsmaller{\hRrGML(h) = \frac{1}{n} \sum\nolimits_{i=1}^n \ell(h(A,Z)_i, y_i)}
\end{equation}

\paragraph{Risk-transferability.} Defining a notion of expected risk is far from obvious, as the empirical risk $\hRrGML$ is \emph{not} a sum of independent variables. 
Moreover, we need a mechanism to \emph{study large training data $n \to \infty$}, i.e. to ``let'' the training graph grow and hopefully materialize the fact that a bigger graph is supposed to be better for generalization. This is intimately linked to the notion of \textbf{transferability} between sizes \cite{transfer_transferability}. Here we will consider arguably the ``weakest'' notion of transferability necessary to study generalization: transferability of the \emph{expected risk} itself. To do so, we consider a \textbf{collection of graph distributions}
$$
\pP = \{P_n\}_{n \geq 1}
$$
where $P_n = P_{G,Y|n}$ is a shorthand notation for a distribution over graphs and node labels $(G,Y)$ \emph{of size $n$}. We will call $\pP$ a \textbf{random graph model}. Then, we denote by $\RrGMLn$ the expectation of $\hRrGML$ over graphs of size $n$:
\begin{equation}\label{eq:expectedriskGMLn}
    \RrGMLn(h) = \EE_{(G,Y) \sim P_n} \hRrGML(h)
\end{equation}
Interestingly, we show in Appendix \ref{app:equivariance} that, as the distribution $P_n$ is exchangeable\footnote{i.e. every permutation of a graph have the same probability}, then $\RrGMLn = \EE_{(G,Y) \sim P_n} \ell(h(A,Z)_1, y_1)$, that is, by equivariance we measure the error \emph{over one node} (arbitrarily, the first one).

Then, we naturally materialize transferability by having $\RrGMLn \approx \RrGML^{(m)}$ for all $n,m$ large enough. 
By properties of Cauchy sequences, this is equivalent to having the risk \emph{converge}\footnote{In \cite{transfer_gmn}, the authors show in their case that $\sup_{m \geq n} \abs{\RrGMLn(h) - \RrGML^{(m)}(h)}$ vanishes when $n \to \infty$, which is equivalent to the existence of a limit quantity.} (pointwise) when $n \to \infty$. Since this convergence is not \emph{a priori} guaranteed for all functions $h \in \Hheq$, we will thus restrict the function space to add this constraint.
\begin{definition}\label{def:transfer}
Given a random graph model $\pP$ and a loss $\ell$, the set of \textbf{risk-transferable functions} is defined as
\begin{align}\label{eq:Feqinfty}
    &\Hhtrans := \left\lbrace h \in \Hheq ~|~ (\RrGMLn(h))_{n \geq 1} \text{ converges}\right\rbrace
\end{align}
For $h \in \Hhtrans$, we denote by $\RrGML(h)\in \RR_+$ the limit risk.
\end{definition}

As mentioned in the literature, this is arguably the ``weakest'' notion of transferability, where we only need the \emph{expected risk} to transfer. 
Note that the set of risk-transferable functions might well be empty, depending on the random graph model!
In particular, the limit quantity $\RrGML(h)$ has no reason to be a ``simple'' mapping of the function $h$. It is just a \emph{scalar} in $\RR_+$ defined as a limit. 
We will see later examples where $\RrGML(h)$ has a friendly closed-form expression, and relate risk-transferability to existing convergence results \cite{Levie2021, Ruiz2020b, Keriven2020a}. We leave studying the links with metric-based transferability results \cite{Levie2023, transfer_transferability} for future work. 

Our choice of notation is deliberate: we now consider that $\RrGML$ plays the role of ``true'' risk, since it will approximate all expected risks over large enough graphs, and thus risk-transferable functions $\Hhtrans$ to be the ``base'' space to seek out the optimal prediction function $h^\star$:
\begin{equation*}
\RrGML^\star = \inf_{h \in \Hhtrans} \RrGML(h),\quad h^\star = \uargmin{h \in \Hhtrans} \RrGML(h)\, .
\end{equation*}

\paragraph{Excess risk decomposition.} Similar to the function set $\Ff \subset \Ff_{all}$ in ML, we consider prediction models 
$\Hh \subset \Hhtrans$, and derive a new excess risk decomposition: denoting as before $\hat h = \uargmin{h \in \Hh} \hRrGML(h)$ our main estimator, and $\RrGML^\Hh = \inf_{h \in \Hh} \RrGML(h)$, an easy computation leads to
\begin{align}
    &
    \RrGML(\hat h) - \RrGML^\star 
    \leq 2\underbrace{\sup_{h \in \Hh}\abs{\RrGMLn(h) - \RrGML(h)}}_{\text{transferability error}} \notag\\& + 2\underbrace{\sup_{h \in \Hh} \abs{\hRrGML(h) - \RrGMLn(h)}}_{\text{estimation error}} + \underbrace{\RrGML^\Hh - \RrGML^\star}_{\text{approximation error}}\label{eq:excessriskdecompositionGML}
\end{align}

The risk decomposition now involves three errors. The estimation error is now between the empirical risk computed on $G$ and its expectation over graphs \emph{of the same size}. 
Like before, the approximation error expresses how much $h_\Hh$ can approximate $h^\star$. 
And the novel term is the \emph{transferability error}, which expresses the convergence of $\RrGMLn(h)$ uniformly over the function set $\Hh$.
At this point, we cannot really go further in the analysis without more specification on $\pP$ or $\Hh$, which we do in the next two sections.

\section{Latent Position Random Graphs}\label{sec:lpm}

We now instantiate the framework of the previous section for Latent Positions Models (LPMs), a general family of random graphs that include many well-known examples. 


\subsection{Background on LPMs}

In an LPM, a graph with $n$ nodes is generated as follows: each node is equipped with an unobserved \emph{latent variable}\footnote{we purposely use the notation $x$ here, for reasons that will be clear later} $x_i \in \Xx$, and for each pairs of nodes, random edges $a_{ij}$ are drawn \emph{independently} conditionally on $(x_i,x_j)$, and potentially on $n$.  
For binary edges, this is a Bernoulli distribution whose parameter is a function $w_n:\Xx \times \Xx \to [0,1]$ often referred to as \emph{connectivity kernel}: 
\begin{equation}
x_i \simiid P_x, \quad a_{ij} \sim \text{Ber}(w_n(x_i,x_j))\, . \label{eq:LPM}
\end{equation}
In the literature, the case of non-random weighted edges $a_{ij} = w_n(x_i,x_j)$ is sometimes considered for simplicity \cite{Cordonnier2023, Keriven2021, Wang2022b}. We will loosely refer to Bernoulli edges as ``random edges'', and weighted edges as ``deterministic edges'', even though the $x_i$ are random.

The dependence in $n$ allows to tune the sparsity of the graph: when $w_n = w$ does not depend on $n$, the generated graphs are dense in expectation, i.e. the expected number of edges is in $\order{n^2}$. 
Otherwise, sparsity is generally introduced either through a multiplicative factor $
    w_n(x_i,x_j) = \alpha_n w(x_i,x_j)
$, such that the expected number of edges is in $\order{\alpha_n n^2}$, 
%
or through a varying \emph{bandwidth} $
    w_n(x_i,x_j) = \tilde w\pa{\frac{x_i-x_j}{t_n}}
$. 
Under some mild hypotheses, the expected number of edges is in $O(n^2 t_n^{d_x})$. 
Classical examples of LPMs include: Erdös-Rényi \cite{Erdos1959}, Stochastic Block Model (SBM) \cite{Holland1983}, random geometric graphs \cite{Penrose2008}, dot-product kernels \cite{Araya2019}, graphons \cite{Lovasz2012}...

\paragraph{Node features and node labels}
Classical LPMs generate only graph structures. 
For node features and labels, a variety of choices exist in the literature, but it remains an active domain \cite{Levie2023}. A popular choice \cite{Keriven2023} is to draw node features and labels independently according to a distribution \emph{conditioned} on individual latent variables for each node:
\begin{equation}\label{eq:nodefeatures}
    z_i \sim P_z(\cdot~|~ x_i),\quad  y_i \sim P_y(\cdot~|~ x_i)
\end{equation}
This is for instance the case of the popular Contextual SBM \cite{Deshpande2018}. In the literature \cite{Keriven2020a, Ruiz2020b}, a simple example of node features is to assume a function $z_i = f(x_i)$, that typically only yields partial information about $x_i$.

In the rest of the paper, $\pP$ will be an LPM. A distribution $P_n$ now concerns $(A,Z,Y)$, but also the latent variables $X$, even if they are unobserved.

\subsection{Learning on LPMs is harder than classical ML}

All along this section, we have purposely used notations reminiscent of classical ML. 
Indeed, LPMs allow for interesting connections between the two worlds.
Fig.~\ref{fig:LPMs} summarizes the similarities and differences: in ML one observes
$
(x_i,y_i) \simiid P_{xy}
$
and aim to predict $y$ from $x$, and in node-tasks on LPMs, there exist
$
(x_i, y_i, z_i) \simiid P_{xyz}
$, however the $x_i$ are not known, and instead one observes the typically less informative $z_i$, as well as the graph structure $A$, which contains information about \textbf{the whole collection of $x_i$'s}. 

Observe that, given an LPM $\pP$, one can define all the ML risks $\RrML$ over $(x,y)$ (even if $x$ is unobserved), along with the GML risks on $(A,Z)$. Then we have the following.

\begin{figure*}[htbp]
  \centering
  \begin{subfigure}[t]{0.38\textwidth}
    \centering
    \scalebox{0.8}{
      \begin{tikzpicture}[
        node/.style={draw,circle,minimum size=6mm,inner sep=0,align=center},
      ]
        \node[node,fill=blue!40] (x1) at (0,1.5) {$x_1$};
        \node[node,fill=red!40] (y1) at (0,0) {$y_1$};
        \draw (x1) -- (y1);
        \node[node,fill=blue!40] (x2) at (1,1.5) {$x_2$};
        \node[node,fill=red!40] (y2) at (1,0) {$y_2$};
        \draw (x2) -- (y2);
        \node[] (dot) at (2, 1.5) {$.~.~.$};
        \node[node,fill=blue!40] (x) at (3,1.5) {$x_n$};
        \node[node,fill=red!40] (y) at (3,0) {$y_n$};
        \draw (x) -- (y);
      \end{tikzpicture}
    }
    \caption{In ML, one observes a collection of i.i.d. training data $(x_i,y_i)$.} 
    \label{fig:sub:left}
  \end{subfigure}\hfill
  \begin{subfigure}[t]{0.58\textwidth}
    \centering
    \scalebox{0.8}{
      \begin{tikzpicture}[
        node/.style={draw,circle,minimum size=6mm,inner sep=0,align=center},
        ann/.style={font=\small, align=center}
      ]
        \node[node,fill=blue!40] (x) at (0,1.5) {$x_1$};
        \node[node,fill=red!40] (y) at (-.4,0) {$y_1$};
        \node[node,fill=green!40] (z) at (.4,0) {$z_1$};
        \node[] (dot) at (1, 1.5) {$.~.~.$};
        \node[node,fill=blue!40] (xn) at (2,1.5) {$x_n$};
        \node[node,fill=red!40] (yn) at (1.6,0) {$y_n$};
        \node[node,fill=green!40] (zn) at (2.4,0) {$z_n$};
        \node[node,fill=orange!40] (a) at (3.5,0) {$A$};
        \draw (x) -- (y);
        \draw (x) -- (z);
        \draw (xn) -- (yn);
        \draw (xn) -- (zn);

        \node[draw,thick,blue, opacity=1, ellipse, minimum width=100pt, minimum height=30pt] (E) at (1,1.5) {};
        \draw[color=blue, thick] (a) -- (E.south east);

        \node[ann] (obs) at (5.2,1.5) {Unobserved};
        \node[ann] (obs) at (5.2,0) { Observed};
      \end{tikzpicture}
    }
    \caption{In graph ML on LPMs, the $x_i$ are not observed, but only partially known through the node features $z_i$'s and the graph structure $A$.}
    \label{fig:sub:right}
  \end{subfigure}
  \caption{Statistical representation of ML (left) and graph ML on LPMs (right).}
  \label{fig:LPMs}
  \vspace{-10pt}
\end{figure*}

\begin{proposition}
\label{prop:Bayes}
    For all $h \in \Hheq$ and all $n$,
    \begin{equation}\label{eq:bayesrisknGML}
        \RrGMLn(h) \geq \RrML^\star
    \end{equation}
    In particular, for all $h \in \Hhtrans$ we have $\RrGML(h) \geq \RrML^\star$, such that $\RrGML^\star \geq\RrML^\star$. 
\end{proposition}
That is, since the $x_i$'s are unknown, graph ML is provably ``more difficult'' than regular ML, in the sense that the optimal risk will always be higher.
An immediate but interesting corollary is the following.
\begin{corollary}\label{cor:optimalh_Xknown}
    Let $\pP$ be a LPM \textbf{for which the latent variables are observed}: $Z=X$. Then $\RrGML^\star =\RrML^\star$, and if the Bayes predictor $f^\star$ exists,
    $
        h^\star(A,X) = [f^\star(x_i)]_{i=1}^n
    $.
\end{corollary}

In other words, \emph{when the $x_i$ are known}, the optimal $h^\star$ returns $f^\star(x_i)$ independently for each node and \textbf{completely ignores the graph structure}! Which intuitively makes sense, as it contains no additional information.

\subsection{Bounding the excess risk on LPMs}\label{sec:learning_lpm}

As we will now see, the framework of LPMs allows to push significantly further the analysis of Section \ref{sec:gml_risk}. 

\paragraph{Latent-convergence and risk-transferability.} Recall that a central tool of our framework is the notion of \emph{risk-transferable functions} $\Hhtrans$ (Def.~\ref{def:transfer}),
While $\Hhtrans$ is hard to express in general, for LPMs a substantial simplification can be brought by a more natural notion of convergence towards functions \emph{on the latent space}, which is actually one of the main approach for GNN transferability in the literature \cite{Keriven2020a, Maskey2022, Ruiz2020b, Levie2021}. We call this notion \textbf{latent-convergence}.

\begin{definition}[Latent-convergent set of functions]\label{def:Pconv}
    Let $h \in \Hheq$. If there exists $f \in \Ff_{all}$ such that
    \begin{equation}\label{eq:Pconv}
        \EE_{P_n} d_\oYy(h(A,Z)_1, f(x_1)) \leq r_n \xrightarrow[n \to \infty]{} 0
    \end{equation}
    we say that $h$ is \textbf{latent-convergent} towards $f$ with rate $r_n$.
    For a set of latent-convergent functions $\Hh \subset \Hheq$, we denote by $\Ff_{\Hh}$ the set of limit functions: 
    \begin{equation}\label{eq:Pconvset}
        \Ff_{\Hh} = \{f ~|~ \exists h \in \Hh \text{ that is latent-convergent to $f$}\}
    \end{equation}
    We say that \textbf{$\Hh$ is latent-convergent towards $\Ff_{\Hh}$}. If all functions in $\Hh$ are latent-convergent with a (maximum) rate of $r^{(\Hh)}_n$, we say that $\Hh$ is latent-convergent with rate $r^{(\Hh)}_n$.
\end{definition}
That is, $h$ is latent-convergent if, in expectation, the prediction $h(A,Z)_1$ converges\footnote{Some works \cite{convergence_ign,Wang2022b} express convergence in probability instead. Since all considered quantities are usually bounded almost surely, this is an equivalent notion, but the rate of convergence is less direct.} to a function \emph{of the latent variable} $x_1$ (which we recall is unobserved). For instance, if $z_i = f(x_i)$, then the identity function $h(A,Z) = Z$ is immediately latent-convergent towards $f$ with rate $0$. But of course the interest lies in more complex functions. Note that, again by \emph{exchangeability} and the fact that we are considering an expectation, the index $1$ in Def.~\ref{def:Pconv} is arbitrary, and could be replaced by any $i$. 

The next theorem linking latent-convergence and risk-transferability is key.
\begin{theorem}\label{thm:finfty}
    Suppose that the loss satisfies Ass.~\ref{ass:loss}. 
    Let $h \in \Hheq$ be latent-convergent to $f$. Then $h \in \Hhtrans$, and 
    \begin{equation}\label{eq:equalrisk}
        \RrGML(h) = \RrML(f)
    \end{equation} 
\end{theorem}
In other words, a latent-convergent function is \emph{automatically} risk-transferable, and the limit risk $\RrGML(h)$ \textbf{is the ML risk} $\RrML(f)$ \textbf{of the limit}. 
Of course, latent-convergence is just one tool: generally, the set of risk-transferable functions may be strictly larger than that of latent-convergent functions. 
\begin{proposition}
    \label{prop:csbm}
    There exist an LPM $\pP$ and loss $\ell$ satisfying Ass.~\ref{ass:loss} such that there exists a risk-transferable function $h \in \Hhtrans$ that is \textbf{not} latent-convergent.
\end{proposition}

Nevertheless, latent-convergence provides a practical way to prove that functions are risk-transferable, and more importantly it provides an actual, simple expression for the limit risk $\RrGML$. We will see in the next sections that it allows to entirely revisit the risk decomposition \eqref{eq:excessriskdecompositionGML}.

\subsubsection{Transferability error}

The transferability error in \eqref{eq:excessriskdecompositionGML} can be easily deduced from latent-convergence. 
\begin{proposition}
    \label{prop:transferconv}
Assume that $\Hh$ is latent-convergent towards $\Ff_\Hh$ with rate $r^{(\Hh)}_n$, and that Ass.~\ref{ass:loss} holds on the loss. 
Then, \textbf{the transferability error in \eqref{eq:excessriskdecompositionGML} is bounded by $L r^{(\Hh)}_n$}, 
where $L = L_\ell$ when the loss is Lipschitz and $L \lesssim \sup_{f \in \Ff_\Hh} \EE_x \norm{f(x)} + D_\Yy$ when it is the square loss.
\end{proposition}

\subsubsection{Approximation error}\label{sec:approxerrorLPM}

With latent-convergence, we are able to relate the approximation error \eqref{eq:excessriskdecompositionGML} to more classical notions of approximation for functions on $\Xx$. As mentioned before, for deep neural networks the basic property is \emph{universality}, which we now formally define. We equip $\Ff_{all}$ with the following metric:
\begin{equation*}
    d_{\oYy,\infty}(f, g) = \sup_x d_{\oYy}(f(x), g(x))
\end{equation*}

\begin{definition}
    [Universality]\label{def:univ}
    Let $\fF = \{\Ff^R\}_{R>0}$ be a collection of function sets $\Ff^R \subset \Ff_{all}$ parameterized by some \textbf{capacity} $R$, let $\Cc \subset \Ff_{all}$ be another set of functions. We say that $\fF$ is \textbf{universal} for $\Cc$ if: for any $f \in \Cc$ and any $\epsilon>0$, there exists $R_\epsilon$ and $f_\epsilon \in \Ff^{R_\epsilon}$ such that
    $
        d_{\oYy,\infty}(f_\epsilon, f) \leq \epsilon
    $.
\end{definition}
Typically, $\Ff^R = \{f_\theta~|~ \Omega(\theta)\leq R\}$ is a set of parametric functions, where $\Omega$ expresses the dimensionality and maximum amplitude of $\theta$. 
For instance, it is known that deep neural nets are universal for continuous functions $\Cc$. 
%
The following theorem relates universality for functions on $\Xx$ with the approximation error \eqref{eq:excessriskdecompositionGML}. 


\begin{theorem}[Approximation error of universal models]\label{thm:univpgconv}
    Let $\pP$ be a LPM, 
    and $\Hh^R\subset \Hhtrans$ be a family of models. Assume that for each $R$, the set $\Hh^R$ is latent-convergent towards $\Ff_{\Hh^R}$, and that there is a set of functions $\Cc$ such that the collection $\fF = \{\Ff_{\Hh^R}\}_{R>0}$ is \textbf{universal} for $\Cc$. Suppose that the loss satisfies Ass.~\ref{ass:loss}.
    %
    If either:
    \begin{enumerate}
        \item $h^\star$ exists and is latent-convergent towards some $f \in \Cc$ 
        \item a Bayes estimator $f^\star$ exists and is in $\Cc$. 
    \end{enumerate}
    Then for all $\epsilon>0$, there is $R_\epsilon$ such that, taking $\Hh = \Hh^{R_\epsilon}$, 
    \textbf{the approximation error \eqref{eq:excessriskdecompositionGML} is bounded by $L\epsilon$}, 
    where $L = L_\ell$ if the loss is $L_\ell$-Lipschitz, or $L \lesssim \sqrt{\RrGML(h^\star)} + D_\Yy$ if the loss is the square loss.
\end{theorem}

In other words, if either the optimal prediction function $h^\star$ is latent-convergent towards a function in $\Cc$, or the Bayes predictor $\fML^\star$ belong to $\Cc$, then the approximation error vanishes by leveraging the universality of $\{\Ff_{\Hh^R}\}_{R>0}$ in $\Cc$.
%

\subsubsection{Estimation error by covering numbers}\label{sec:estimerrorLPM}


We are left to deal with the estimation error in \eqref{eq:excessriskdecompositionGML}. Here we develop a generic approach for LPMs, based on bounded differences, a variant of Mc Diarmid's inequality, and covering numbers to obtain a uniform bound. It is known that covering numbers are generally suboptimal in ML \cite{Bach2021}, however more advanced approaches such as Rademacher complexities fall short here. In particular, in our case the \emph{random graph varies} in the risk, and therefore we cannot applied variants that consider a fixed dependency graphs between variables \cite{morris_node_generalization}. We leave more optimal approaches for future work.

\begin{table*}
    \centering
    \small
    \renewcommand{\arraystretch}{1.3} 
    \begin{tabular}{|l||c|c|c|c|}
        \toprule
        \multicolumn{1}{|c||}{\textbf{Example}} & \textbf{Graph model} & \textbf{Transfer. error} & \textbf{Estim. error} & \textbf{Universality} \\ \midrule[\heavyrulewidth]
        \multirow{2}{*}{Normalized adjacency} & $a_{ij} \sim \text{Ber}(\alpha_n w(x_i,x_j))$ & $1/\sqrt{\alpha_n n}$ &  $\frac{\sqrt{\log n}}{\min(n \alpha_n, \sqrt{n})}$& .\\
        & $a_{ij} = w(x_i,x_j)$ & $1/\sqrt{n}$ & $\sqrt{(\log n) / n}$ &  Yes with PE \\ \midrule
        \multirow{2}{*}{GCN}  & $a_{ij} \sim \text{Ber}(\alpha_n w(x_i,x_j))$ & $1/\sqrt{\alpha_n n}$ &$\frac{\sqrt{\log n}}{\min(n \alpha_n, \sqrt{n})}$ & .\\ 
        & $a_{ij} = w(x_i,x_j)$ & $1/\sqrt{n}$ & $\sqrt{(\log n) / n}$ &.\\\midrule
        GAT & $a_{ij} = w(x_i,x_j)$ & $\sqrt{(\log n) / n}$ & . &.\\\midrule
        Generalized mean aggregation  & $a_{ij} = w(x_i,x_j)$ & $((\log n) / n)^{\alpha_g \alpha_F^L/2}$ & . & .\\\midrule
        Max. aggregation  & $a_{ij} = w(x_i,x_j)$ & $\pa{\frac{\log n}{n}}^{1/d}$& . & .\\\midrule
        IGN & $a_{ij} = w(x_i,x_j)$ & $\frac{1}{n^{1/4}}$& . & . \\ \midrule
        Manifold NN & $a_{ij} = \tilde w_n\pa{\frac{\norm{x_i-x_j}}{t_n}}$ & \multicolumn{2}{c|}{$\order{\pa{\frac{\log n}{n}}^{1/d_\Mm}}$} & . \\ \bottomrule 
    \end{tabular}
    \caption{Summary of latent-convergence rates for several GNNs, and estimation error for two basic examples. In each case, the graph structure must satisfy some assumptions on the kernel $w$, and rates $\alpha_n\gtrsim (\log n) / n$ or $t_n$, see App.~\ref{app:gnn} and original paper in each case.}\label{tab:example}
    \vspace{-10pt}
\end{table*}

We consider $\Hh \subset \Hhtrans$, and fix a graph size $n$ in this section. 
Our main bounded difference assumption deviates slightly from the usual application of McDiarmid's inequality, in the sense that we assume that it holds over a set of graphs $\Gg_n$: this will be useful in some examples. 

\begin{assump}\label{ass:stability_estim_error}
     Assume that, for all $n$, there is a set of graphs $\Gg_n$ of size $n$ such that, for all $h \in \Hh$ and $(A,Z), (A', Z') \in \Gg_n$: denoting by $E_h = \sum_{k=1}^n d_\oYy(h(A,Z)_k, h(A', Z)_k)$, we have
    \begin{itemize}
        \item if $Z=Z'$ and $A, A'$ differ only on one entry, $E_h \leq \Delta^a_n$
        \item if $Z=Z'$ and $A,A'$ differ on one line and column (since they are symmetric), $E_h \leq \Delta^A_n$
        \item if $A=A'$ and $Z,Z'$ differ only on one row, $E_h \leq \Delta^z_n$
    \end{itemize}    
\end{assump}

Moreover, we will need the following global bounds on the loss, which is generally just a technicality. 
\begin{assump}\label{ass:lossbounded}
    Considering the same set $\Gg_n$ as in Ass.~\ref{ass:stability_estim_error}, assume that $\sup_{h \in \Hh} \sup_{y \in \Yy}\max_i \abs{\ell(h(A,Z)_i, y)} \leq c_n $ if $(A,Z) \in \Gg_n$, or $C_n$ otherwise.
\end{assump}



To obtain a uniform bound, we proceed by covering numbers. Define the metric over $\Hheq$:
\begin{equation*}
    d_{\oYy, n}(h,h') = \sup_{(A,Z)\in \text{supp}(P_n)} \max_{i} d_{\oYy}(h(A,Z)_i, h'(A,Z)_i)
\end{equation*}
where $\text{supp}(P_n)$ is the support of $P_n$. We denote by $\Nn(\Hh, \epsilon, d_{\oYy,n})$ the covering numbers of $\Hh$, i.e., the number of balls of radius $\epsilon$ for the metric $d_{\oYy, n}$ to cover it. 

\begin{theorem}[Estimation error through covering numbers.]\label{thm:stabilitycovering}
    Suppose that Ass.~\ref{ass:loss}, \ref{ass:stability_estim_error} and \ref{ass:lossbounded} hold. Denote by $p_n = P_n(\Gg_n^c)$, and $\Nn = \Nn(\Hh, E_n/\sqrt{n}, d_{\oYy,n})$, where $E_n$ is defined below. With probability at least $1-\delta - \Nn p_n$, \textbf{the estimation error \eqref{eq:excessriskdecompositionGML}} satisfies
    \begin{align}
        &\sup_{h\in\Hh}\abs{\hRrGML(h) - \RrGMLn(h)} \lesssim (D_n + n L \tilde \Delta^a_n + C_n) p_n \notag \\&\quad + (D_n + \sqrt{n} L \tilde \Delta^a_n) \sqrt{\tfrac{\log(\Nn/\delta)}{n}}
    \end{align}
    where $D_n = L_\ell(\Delta^A_n + \Delta^z_n) + c_n$, $L=L_\ell$ and $E_n = D_n/L_\ell$ for Lipschitz bounded loss, or $D_n = (c_n + D_\Yy)(\Delta^A_n + \Delta^z_n + D_\Yy)$, $L = c_n + D_\Yy$ and $E_n = D_n/(D_\Yy + C_n)$ for the square loss; and $\tilde \Delta^a_n = \Delta^a_n$ for random edges, or $\tilde \Delta^a_n=0$ for deterministic edges.
\end{theorem}

We can often design $\Gg_n$ to have $p_n$ negligible in front of the other terms, and for many models we even have $p_n=0$. Moreover, we will generally have $D_n = \order{1}$, and $\Delta^a_n = o(1/\sqrt{n})$. For parametric functions, the covering numbers generally yields $\log(\Nn) \propto d_\Theta \log(n)$, where $d_\Theta$ is the dimensionality of the parameters. 
Thus, we will generally obtain a global rate in $O(\sqrt{(d_\Theta \log n)/n})$ for dense LPMs. We may also obtain a slower rate for sparser graphs, see next section. 



\section{Application to GNNs}\label{sec:gnn}


In this section, we apply the previous framework to examples of GNNs. We mainly report \emph{known} results of transferability and universality, bound the estimation error in certain cases, and leave other bounds for future work. See Tab.~\ref{tab:example}. In Appendix \ref{app:general}, we develop \emph{generic} approaches for any GNN, that allow us to recover the results presented here.



\subsection{Latent-convergence and transferability error}

As mentioned before, convergence to limit objects has been a popular approach to study GNN transferability \cite{Keriven2020a,Levie2021,Ruiz2020b,Cordonnier2023}. Often, they are latent-convergent towards ``continuous'' GNNs (cGNN) that propagate functions on the latent space with a ``continuous message-passing''.
For instance, in many cases the aggregation can be modelled as multiplication with a graph matrix $S=S(A)$: 
\begin{equation}\label{eq:SGNN}
    Z^{(k)} = \rho\pa{Z^{(k-1)}\theta_0^{(k)} + S Z^{(k-1)}\theta_1^{(k)} + 1_n {b^{(k)}}^\top}
\end{equation}
with weights $\theta^{(k)}_p$, bias $b^{(k)}$, and non-linearities $\rho$. 
Then, $S$ might ``converge'' to a continuous operator $\vS$, such that the GNN converges to a cGNN that propagates $f^{(k)} = \rho_k((\theta_0^{(k)})^\top f^{(k-1)} + (\theta_1^{(k)})^\top \vS f^{(k-1)} + b^{(k)})$. Latent-convergence has also been shown in more general examples of message-passing \cite{Cordonnier2023}. 



We report known results of latent-convergence here, see Tab.~\ref{tab:example}. We assume that the node features are directly $z_i = f^{(0)}(x_i)$ for simplicity. All details can be found in App. \ref{app:gnn}.

%

\begin{itemize}
    \item \textbf{Normalized sum aggregation}: here $S = A/(\alpha_n n)$, assuming $\alpha_n$ is known for simplicity. The cGNN uses the operator $\vS f = \int w(x,y)f(y)dP_x(y)$. Convergence results are from \cite{Keriven2021}.
    \item \textbf{GCN} \cite{Kipf2017}: the classical GCN. The convergence results are similar, with an adapted operator \cite{Keriven2020a}.
    \item \textbf{GAT} \cite{Velickovic2018}: uses attention coefficient on the edges. Latent-convergence holds with deterministic edges, towards an adapted operator with modified attention kernel \cite{Cordonnier2023}.
    \item \textbf{Generalized mean aggregation} \cite{general_aggregation_kortsevely}: a less-classical example, with aggregation as a ``generalized mean'', that includes e.g. geometric mean. Convergence results are from \cite{Cordonnier2023}, and use some special rate that depends on coefficients $\alpha_g,\alpha_F$ and number of layers $L$. See details in appendix.
    \item  \textbf{Max. aggregation} \cite{Hamilton2017}: the classical, albeit not-often-used, aggregation as a maximum over the neighbors. Yields a slow convergence rate that depends on the dimension of the space \cite{Cordonnier2023}
    \item \textbf{IGN} \cite{Maron2019,convergence_ign}: Invariant Graph Network, a GNN that uses high-order invariant linear layers \cite{Maron2019}. Convergence results are from \cite{convergence_ign}.
    \item \textbf{Manifold NN} \cite{Wang2022b}: a specific construction where the kernel has a shrinking bandwidth, and the $x_i$ are distributed over a $d_\Mm$-dimensional manifold. The limit function uses the Laplace-Beltrami operator. Latent-convergence is shown by \cite{Chew2022a}, but here we report results by \cite{Wang_manifold_generalization}, who bound a quantity that is roughly equivalent to both estimation and transferability error at once. The kernel $\tilde w_n$ is either a Gaussian kernel or a random geometric graph, with some additional normalization in $n$. 
\end{itemize}

\subsection{Estimation error for two examples}

Here we derive the estimation error \eqref{eq:excessriskdecompositionGML} for the first two examples above: aggregation with normalized adjacency matrix or GCN matrix. We consider Bernoulli edges $a_{ij} = \text{Ber}(\alpha_n w(x_i, x_j))$ with $\alpha_n \gtrsim (\log n) / n$, assume that $w$ is Lipschitz, and for GCN $\inf_y \int w(y,x)dP_x(x) \geq c_{\min}>0$ \cite{Keriven2020a}. For weighted edges $a_{ij} = w(x_i,x_j)$, we get the same result taking $\alpha_n=1$.

As detailed in Appendix \ref{app:gnn}, we show in this case that for any $M>0$, there is a constant $C_M$ such that Ass. \ref{ass:stability_estim_error} is satisfied with  $p_n = n^{-M}$, $\Delta^A_n, \Delta^z_n$ and $c_n$ in $\order{1}$, $\Delta^a_n = \order{1/(\alpha_n n)}$, and $C_n = \order{1/\alpha_n^L}$. Here all multiplicative constants depend polynomially on $\norm{\theta^{(k)}_p}, \norm{b^{(k)}}$ and $C_M$, and $c_{\min}$ in the case of GCN.
By computing the covering numbers, we get the following result.
\begin{proposition}
    Under the assumptions detailed above,
    \textbf{the estimation error is in} $\order{R_\Theta\frac{\sqrt{\log n}}{\min(n \alpha_n, \sqrt{n})}}$
    where $R_\Theta$ is a constant that depends polynomially on the maximum amplitude of the parameters and on $c_{\min}$ in the case of GCN, and linearly on the number of parameters.
\end{proposition}

\subsection{A complete example with universality}\label{sec:univ}

Finally, we give a final example, for which universality has been proven \cite{Keriven2021}, if we add certain Positional Encodings (PE) as input. 
Here we assume that we have non-random weighted edges $a_{ij} = w(x_i, x_j)$, and consider the first example above, that of GNNs \eqref{eq:SGNN} with graph operator $S=A/n$. Moreover, we assume that there is no node features (or ignore them) and consider as input the PEs defined in \cite{Keriven2021}
$
    Z^{(0)} = \tfrac{1}{n} \sum_i f_{\theta'}(A_{:,i})
$, 
where $f_{\theta'}:\RR\to \RR$ is an MLP applied element-wise. This is basically a Deepset architecture applied to the rows of $A$ to preserve permutation-equivariance.

\begin{remark}
    These PEs 
    only makes sense if $A$ effectively contains weighted edges (otherwise $Z^{(0)}$ is just proportional to the degrees). In the more realistic case of Bernoulli edges, universality can still be obtained by using a well-chosen \textbf{low-rank approximation of $A$} instead \cite{convergence_ign, Keriven2023}. However this leads to verbose and complex derivations to bound the estimation error, which is out of scope of this paper. We chose the weighted edges case for simplicity.
\end{remark}

\paragraph{Transferability.} As expected, the whole architecture is latent-convergent towards a cGNN with kernel operator $\vS f = \int w(\cdot, x)f(x)dP_x(x)$, and input function corresponding to PEs
$
    f^{(0)} = \int f_{\theta'}(w(\cdot, x))dP_x(x)
$
As proven in \cite{Keriven2021}, the rate is in $\order{1/\sqrt{n}}$. The transferability error follows.

\paragraph{Estimation error.} Proceeding as above but for non-random edges, we obtain that the estimation error is in $\order{R_\Theta\sqrt{\frac{\log n}{n}}}$, where $R_\Theta$ also depends on the maximum amplitude and dimension of the parameters of the PE $f_{\gamma}$.

\paragraph{Universality.} Finally, the family of limit functions $\fF = \{\Ff_{\Hh^R}\}_{R\geq 0}$, where $R$ bounds the amplitude and dimension of the parameters of both GNN and PEs, has been proven to be universal in the sense of Section \ref{sec:univ} in the following cases (see Appendix \ref{app:univ} for details):
\begin{itemize}
\item \textbf{(Most) SBMs}, under some invertibility condition on the probability matrix;
\item \textbf{Additive kernel} $w(x,x')=v(u(x)+u(x'))$, when $u,v$ are continuous and injective; 
%
\item \textbf{1D radial kernel}: $w(x,x') = v(|x-x'|)$ with continuous and injective $v$: universality among symmetric functions if $P_x$ is symmetric, or all functions otherwise;
\item \textbf{Spherical kernels}: under some injectivity condition on the spherical harmonics of the kernel.
\end{itemize}

\section{Conclusion and outlooks}

We developed a statistical framework for inductive learning from a single large graph, aiming to get as close as possible to classical ML theory. Without surprise, it heavily relies on the notion of transferability, which we formulated on the expected risk itself. On LPMs random graphs, we explored more connections between GML and ML and their relative hardness. Drawing on existing notions of transferability as convergence, we showed that the GML risk can be expressed as a regular ML risk. Finally, we showed that many existing GNN transferability results fit into this framework. 

\paragraph{Outlooks.} Since the primary contribution of this paper is a theoretical framework, we expect that it could be applied to a variety of different situations. First and foremost, we related risk-transferability to convergence results, but another main approach for transferability relies on metrics and regularity \cite{transfer_transferability, Levie2023}. Second, there are missing results in Tab.~\ref{tab:example}. Finally, not much is known about random graph models that are very different from LPMs or graphons: e.g. scale-free models \cite{Barabasi1999}. Deriving conditions for risk-transferability on these models is a major avenue for future work.

\newpage

\subsection*{Acknowledgments}

The authors acknowledge the fundings of France 2030, PEPR IA, ANR-23-PEIA-0008 and European Union
ERC-2024-STG-101163069 MALAGA.

\subsection*{AI use statement}

LLMs were used only for basic typo checks.

{\small
\bibliography{library}
\bibliographystyle{icml2026}
}

\newpage
\appendix
\onecolumn

\section*{Appendix}

This Appendix is organized as follows: in App \ref{app:proofs}, we provide all proofs for the general results on the learning framework. In App \ref{app:gnn}, we describe the GNN examples contained in the paper and prove the related results, or point to references that do. In App \ref{app:general} (seen as a ``bonus''), we describe general methodologies to prove latent-convergence and bound the estimation error for GNNs, first for ``operator'' GNNs \eqref{eq:SGNN} then slightly more general message-passing. The latter is a reframing of the results by Cordonnier et al. \cite{Cordonnier2023}.

\section{General theory: proofs}\label{app:proofs}
In this section, we provide all proofs on the general theory and LPM random graphs.

\subsection{Exchangeability and expected risk}\label{app:equivariance}

In Section \ref{sec:newrisk} we mentioned the fact that the expection of the empirical risk $\hRrGML$ can be written as an expectation over \emph{one} node:
$$
\RrGMLn(h) = \EE_{P_n} \frac{1}{n} \sum_{i=1}^{n} \ell(h(A,Z)_i, y_i) = \EE_{P_n} \ell(h( A, Z)_1, y_1)
$$
when the distribution of graph was \emph{exchangeable}, as is generally the case. Denote by $\sigma \cdot G = (\sigma A \sigma^\top, \sigma Z)$ the permutation of a graph by a permutation matrix $\sigma$.
\begin{definition}\label{def:exchangeable}
  A graph/label distribution $P$ over $(G,Y)$ is said to be \textbf{exchangeable} if: for any $n$, any $(G,Y)$ and permutation matrix $\sigma$ of size $n$, $(\sigma \cdot G, \sigma Y)$ has the same law as $(G,Y)$.
\end{definition}
We have the following proposition.
\begin{proposition}\label{prop:equivariance}
  Let $P$ be a graph distribution (potentially over all graph sizes). If $P$ is exchangeable, then for any $h \in \Hh_{all}$
  \begin{equation}\label{eq:expectedriskGML}
    \RrGML(h) = \EE_{(G,Y) \sim P} \ell(h(A,Z)_1, y_1)
  \end{equation}
\end{proposition}

\begin{proof}
  Considering that the graph size is random, we can decompose $\EE_{G,Y} = \EE_n \EE_{G,Y|n}$, where the second expectation is over the distribution of graphs of size $n$. Taking $\sigma_i$ the permutation that exchange $i$ and $1$, since $(\sigma_i A \sigma_i^\top, \sigma_i Z, \sigma_i Y)$ has the same distribution as $(A,Z,Y)$ and $h$ is equivariant,
  \begin{align*}
    \RrGML(h) &= \EE_{G,Y} \frac{1}{|G|} \sum_{i=1}^{|G|} \ell(h(A,Z)_i, y_i) = \EE_n \frac{1}{n} \sum_{i=1}^n \EE_{A,Z,Y|n}\ell(h(A,Z)_i, y_i) \\
    &= \EE_n \frac{1}{n} \sum_{i=1}^n \EE_{A,Z,Y|n}\ell(h(\sigma_i A \sigma_i^\top,\sigma_iZ)_i, [\sigma_i Y]_i) = \EE_n \frac{1}{n} \sum_{i=1}^n \EE_{A,Z,Y|n}\ell([\sigma_i h( A, Z)]_i, [\sigma_i Y]_i) \\
    &= \EE_n \frac{1}{n} \sum_{i=1}^n \EE_{A,Z,Y|n}\ell(h( A, Z)_1, y_1) = \EE_{G,Y}\ell(h( A, Z)_1, y_1)
  \end{align*}
\end{proof}

Similarly, in the definition of latent-convergence Def.~\ref{def:Pconv}, we have:
\begin{align*}
  \EE_{P_n} d_\oYy(h(A,Z)_1, f(x_1)) = \EE_{P_n} d_\oYy(h(A,Z)_i, f(x_i))
\end{align*}
for all $i$, where $P_n$ now involves $X$ in an exchangeable manner as well.

\subsection{Proof of Prop.~\ref{prop:Bayes}}

\begin{proof}

  Given a distribution $(x,y)\sim P_{xy}$, it is known \cite{Bach2021} that the Bayes risk is equal to 
  $$
  \RrML^\star = \EE_{x} \inf_{\hat y \in \oYy} [\EE_{y|x}\ell(\hat y, y)] 
  $$

  Then, on an LPM, since $y_1$ only depends on $x_1$,
  \begin{align*}
    \RrGMLn(h) &= \EE_{P_n} \ell(h(A,Z)_1, y_1) = \EE_{X} \EE_{A,Z | X} [\EE_{y_1|x_1}\ell(h(A,Z)_1, y_1)] \\
    &\geq \EE_{X} \EE_{A,Z | X} \inf_{\hat y \in \oYy} [\EE_{y_1|x_1}\ell(\hat y, y_1)] \\
    &= \EE_{x_1} \inf_{\hat y \in \oYy} [\EE_{y_1|x_1}\ell(\hat y, y_1)] = \RrML^\star
  \end{align*}
\end{proof}

\subsection{Proof of Thm.~\ref{thm:finfty}}
\begin{proof}
  When the loss is Lipschitz, the proof is immediate by remarking that for any $n$, $\EE_{x,y} \ell(f(x), y) = \EE_{P_n} \ell(f(x_1), y_1)$ and therefore
  \begin{align*}
    \abs{\RrGMLn(h) - \EE \ell(f(x), y)} &= \abs{\EE_{P_n} \pa{\ell(h(A,Z)_1, y_1) - \ell(f(x_1), y_1)}} \\
    &\leq L_\ell \EE_{P_n} d_\oYy(h(A,Z)_1, f(x_1)) \leq L_\ell r_n
  \end{align*}
  When the loss is the square loss, remark that
  $$
  \abs{\EE_{P_n} \pa{\ell(h(A,Z)_1, y_1) - \ell(f(x_1), y_1)}} \leq \pa{\EE\norm{h(A,Z)_1} + \EE \norm{f(x)} + 2D_\Yy}r_n
  $$
  The only quantity that depends on $n$ in the multiplicative constant is $\EE\norm{h(A,Z)_1}$, and
  \begin{equation*}
    \EE\norm{h(A,Z)_1} \leq \EE\norm{h(A,Z)_1 - y_1} + D_\Yy \leq \sqrt{\RrGMLn(h)} + D_\Yy
  \end{equation*}
  Since $\RrGMLn(h)$ converges, it is uniformly bounded in $n$, which concludes the proof.
\end{proof}

\subsection{Prof of Prop.~\ref{prop:transferconv}}
\begin{proof}
  The Lipschitz case has already been proved in the proof of Thm.~\ref{thm:finfty}. For the square loss, we recall that we have 
  $$
  \abs{\EE_{P_n} \pa{\ell(h(A,Z)_1, y_1) - \ell(f(x_1), y_1)}} \leq \pa{\EE\norm{h(A,Z)_1} + \EE \norm{f(x)} + 2D_\Yy}r_n
  $$
  and $\EE \norm{h(A,Z)_1} \leq \EE\norm{f(x)} + r_n^{(\Hh)}\lesssim \EE\norm{f(x)}$.
\end{proof}

\subsection{Proof of Thm.~\ref{thm:univpgconv}}
\begin{proof}

  Let $\tilde f \in \Cc$ be either $f^\star$ or $\fML^\star$ depending on hypothesis $1)$ or $2)$ in the theorem. In both cases, we have $\RrGML(h^\star) \geq \RrML(\tilde f)$: in the first case, this is an equality by Thm.~\ref{thm:finfty}, and in the second, this is shown by Prop.~\ref{prop:Bayes}.
  
  Take $\epsilon>0$. Since $\fF = \{\Ff_{\Hh^R}\}_{R>0}$ is universal, there exists $R_\epsilon$ and $f_\epsilon \in \Ff_{\Hh^{R_\epsilon}}$ such that
  \begin{equation*}
    d_{\oYy, \infty}(f_\epsilon, \tilde f) \leq \epsilon
  \end{equation*}


  Since $\Hh^{R_\epsilon}$ is latent-convergent towards $\Ff_{\Hh^{R_\epsilon}}$, there exists $h_\epsilon \in \Hh^{R_\epsilon}$ that is latent-convergent towards $f_\epsilon$.
  By minimality of $h_{\Hh^{R_\epsilon}}$ and Proposition \ref{thm:finfty}, we have 
  \begin{equation}
    \RrGML(h_{\Hh^{R_\epsilon}}) \leq \RrGML(h_\epsilon) = \RrML(f_\epsilon)
  \end{equation}
  Hence the approximation error satisfies 
  \begin{align*}
    \RrGML(h_{\Hh^{R_\epsilon}}) - \RrGML(h^\star) &\leq \RrML(f_\epsilon) - \RrML(\tilde f) 
  \end{align*}
  If the loss is $L_\ell$ Lipschitz, we immediately have $\RrML(f_\epsilon) - \RrML(\tilde f) \leq L_\ell d_{\oYy, \infty}(f_\epsilon, \tilde f) \leq L_\ell \epsilon$.
  
  If $\ell$ is the square loss and $d_\Yy$ is the Euclidean norm, we have 
  $$
  \RrML(f_\epsilon) - \RrML(\tilde f) \leq \pa{\EE\norm{f_\epsilon(x)} + \EE\norm{\tilde f(x)} + 2D_\Yy} d_{\oYy, \infty}(f_\epsilon, \tilde f) 
  $$
  and 
  \begin{align*}
    \EE\norm{\tilde f(x)} \leq \EE\norm{\tilde f(x) - y} + D_\Yy \leq \sqrt{\RrML(\tilde f)} + D_\Yy \leq \sqrt{\RrGML(h^\star) } + D_\Yy
  \end{align*}
  and $\EE\norm{f_\epsilon(x)} \leq \epsilon + \EE\norm{\tilde f(x)}$, which allows us to conclude.

\end{proof}

\subsection{Proof of Thm.~\ref{thm:stabilitycovering}}

We start with the following Lemma for pointwise concentration of the risk. 
\begin{lemma}\label{lem:stabilitymcdiarmid}
  Let $\Hh$ satisfying Assumption \ref{ass:stability_estim_error} and \ref{ass:lossbounded}. Suppose that the loss satisfies Assumption \ref{ass:loss}. Denote by $p_n = P_n(\Gg_n^c)$ the probability of generating a graph of size $n$ that is not in $\Gg_n$. For any $h \in \Hh$, with probability at least $1-\delta - p_n$, we have
  \begin{equation}\label{eq:estimerrorGML}
    \abs{\hRrGML(h) - \RrGMLn(h)} \lesssim (D_n + n L \tilde \Delta^a_n + C^{(\ell)}_n) p_n + (D_n + \sqrt{n} L \tilde \Delta^a_n) \sqrt{\frac{\log(1/\delta)}{n}}
  \end{equation}
  where $D_n = L_\ell(\Delta^A_n + \Delta^z_n) + c^{(\ell)}_n$ and $L=L_\ell$ for Lipschitz bounded loss, or $D_n = (c^{(\ell)}_n + D_\Yy)(\Delta^A_n + \Delta^z_n + D_\Yy)$ and $L = c_n^{(\ell)} + D_\Yy$ for the square loss; and $\tilde \Delta^a_n = \Delta^a_n$ for random Bernoulli edges, or $\tilde \Delta^a_n=0$ for deterministic edges.
\end{lemma}
\begin{proof}
  Recall that the collection of independent variables involved in $P_n$ are triplets $(x_i, z_i, y_i)$ for $1\leq i\leq n$, and in the case of random edges, Bernoulli variables can be modelled as
  $$
  a_{ij} = 1(U_{ij} \leq w_n(x_i,x_j))
  $$
  for $U_{ij}$, $i<j$ uniform random variables on $[0,1]$ drawn \emph{independently from all others}. Hence, for weigthed edges the empirical risk is a function of $n$ independent random variables, and for random edges, $n + n(n-1)/2$ independent random variables. Our goal is to compute bounded differences and apply a generalized version of McDiarmid's inequality \cite{Combes2024} (Thm.~\ref{thm:mcdiarmid}).

  Under the event that $(A,Z)$ and $(A',Z')$ both belong to $\Gg_n$, when there is an index $j$ such that $(x_j, y_j, z_j)$ varies: the node features change by one component -- ie, $Z$ and $Z'$ are different only on index $j$, same thing for the labels $Y,Y'$, while the adjacency matrix $A'$ varies from $A$ by column and line $j$. Hence 
  \begin{align*}
    &\frac{1}{n}\abs{\sum_i \ell(h(A,Z)_i, y_i) - \sum_i \ell(h(A',Z')_i, y'_i)} \\
    &\leq 
    \frac{1}{n}\Big(\abs{\sum_i \ell(h(A,Z)_i, y_i) - \ell(h(A',Z)_i, y_i)} 
    + \abs{\sum_i \ell(h(A',Z)_i, y_i) - \ell(h(A',Z')_i, y_i)} \\
    &\quad + \abs{\ell(h(A',Z')_j, y_j) - \ell(h(A',Z')_j, y'_j)}\Big)
  \end{align*}

  When the loss is $L_\ell$-Lipschitz, it results in
  \begin{align*}
    &\frac{1}{n}\abs{\sum_i \ell(h(A,Z)_i, y_i) - \sum_i \ell(h(A',Z')_i, y'_i)} \\ &\leq \frac{1}{n}\pa{L_\ell\sum_i d_\oYy(h(A,Z)_i, h(A',Z)_i) 
    + L_\ell\sum_i d_\oYy(h(A',Z)_i, h(A',Z')_i)
    + 2c_n } \\
    &\leq c_i := \frac{1}{n}\pa{L_\ell(\Delta^A_n + \Delta^z_n) + 2c^{(\ell)}_n} \\
  \end{align*}
  In case of random edges, a variation in $U_{ij}$ results in the same manner to a variation of $c'_{ij} := \frac{L_\ell \Delta^a_n}{n}$.
  We compute $\sum_i c_i + \sum_{ij} c'_{ij}$ and $\sqrt{\sum_i c_i^2 + \sum_{ij} (c'_{ij})^2}$ and applying Thm.~\ref{thm:mcdiarmid} we obtain the result.

  When the loss is the square loss and $d_\Yy$ is the Euclidean norm, since $\max_i \norm{h(A,Z)_i} \leq c^{(\ell)}_n + D_\Yy$ for graphs in $\Gg_n$ by Assumption \ref{ass:stability_estim_error},
  \begin{align*}
    &\frac{1}{n}\abs{\sum_i \ell(h(A,Z)_i, y_i) - \sum_i \ell(h(A',Z')_i, y'_i)} \\
    &\leq \frac{1}{n}\Big(2(c^{(\ell)}_n + D_\Yy)\sum_i \norm{h(A,Z)_i - h(A',Z)_i} 
    \\&+ 2(c^{(\ell)}_n + D_\Yy)\sum_i \norm{h(A',Z)_i - h(A',Z')_i} 
    + 4(c^{(\ell)}_n + D_\Yy)D_\Yy \Big) \\
    &\leq c_i := \frac{2(c^{(\ell)}_n + D_\Yy)}{n}\pa{\Delta^A_n + \Delta^z_n + 2D_\Yy} 
  \end{align*}
  In case of random edges, a variation in $U_{ij}$ results in the same manner to a variation of $c'_{ij} := \frac{2(c^{(\ell)}_n + D_Yy) \Delta^a_n}{n}$. The end of the proof is the same.
  
\end{proof}

We now turn to the proof of the Theorem.

\begin{proof}[Proof of Thm.~\ref{thm:stabilitycovering}]
  The proof is very classical, we provide it here for completeness. Let $\epsilon = 1/\sqrt{n}$, and let $h_1, \ldots, h_\Nn$ be an $\epsilon$-covering of $\Hh$. Using a union bound and Lemma \ref{lem:stabilitymcdiarmid}, with probability $1-\delta-\Nn p_n$, for all $h_i$ we have 
  $$
  \abs{\hRrGML(h_i) - \RrGMLn(h_i)} \lesssim (D_n + n L \tilde \Delta^a_n + C_n) p_n + (D_n + \sqrt{n} L \tilde \Delta^a_n) \sqrt{\frac{\log(\Nn/\delta)}{n}}
  $$

  When the loss is Lipschitz, we have $\abs{\hRrGML(h) - \hRrGML(h')}\leq L_\ell d_{\oYy,n}(h,h')$ and similarly for $\RrGMLn$. Thus, for any $h \in \Hh$, considering $h_i$ such that $d_{\oYy,n}(h,h_i) \leq \epsilon$, we have 
  \begin{align*}
    \abs{\hRrGML(h) - \RrGMLn(h)} &\leq \abs{\hRrGML(h) - \hRrGML(h_i)} + \abs{\hRrGML(h_i) - \RrGMLn(h_i)} + \abs{\RrGMLn(h_i) - \RrGMLn(h)} \\
    &\lesssim 2 L_\ell \epsilon + (D_n + n L \tilde \Delta^a_n + C_n^{(\ell)}) p_n + (D_n + \sqrt{n} L \tilde \Delta^a_n) \sqrt{\frac{\log(\Nn/\delta)}{n}}
  \end{align*}
  which completes the proof.

  When the loss is the square loss, we have $\abs{\hRrGML(h) - \hRrGML(h')}\lesssim (C_n^{(\ell)} + D_\Yy) d_{\oYy,n}(h,h')$ and similarly for $\RrGMLn$, the end of the proof is the same.
  
\end{proof}

\section{GNN examples: description and proofs}\label{app:gnn}
In this section, we describe the GNN examples of Section \ref{sec:gnn} and give reference to the associated proofs, either in other papers or in App.~\ref{app:general} where we provide general hypotheses to show latent-convergence and bound the estimation error of generic GNNs.

The random graph model is an LPM with either random edges $a_{ij} \sim \text{Ber}(\alpha_n w(x_i,x_j))$ or weighted edges $a_{ij} = w(x_i,x_j)$ (in which case $\alpha_n=1$ in the results), or in some sections a more specific model. Unless otherwise mentioned, the kernel $w$ is assumed at least Lipschitz-by-part \cite{Keriven2020}, and the input node features are $z_i^{(0)} = f^{(0)}(x_i)$ for some function $f^{(0)}$. Interested readers can refer to the original papers in each case for more details.

\subsection{Sum aggregation}\label{app:ex_adj}

Here the GNN is:
\begin{equation*}
  Z^{(k)} = \rho_k\pa{Z^{(k-1)}\theta_0^{(k)} + S Z^{(k-1)}\theta_1^{(k)} + 1_n {b^{(k)}}^\top}
\end{equation*}
with $S = A/(n\alpha_n)$ normalized by $n \alpha_n$, where $\alpha_n$ is the sparsity level of the kernel $W_n$ (assumed known for simplicity).

The LPM uses either random edges $a_{ij} \sim \text{Ber}(\alpha_n w(x_i,x_j))$ with $\alpha_n \gtrsim \log n/n$, or weighted edges $a_{ij} = w(x_i,x_j)$. The results below are valid in both cases, with $\alpha_n=1$ for weighted edges. 

\subsubsection{latent-convergence and transferability}

Recalling that $z_i^{(0)} = f^{(0)}(x_i)$, the continuous model propagates
$$
f^{(k)} = \rho_k\circ\pa{{\theta_0^{(k)}}^\top f^{(k-1)} + {\theta_1^{(k)}}^\top \vS f^{(k-1)} + b^{(k)}}
$$
with $\vS f = \int w(\cdot, x)dP_x(x)$, and for multivariate functions $\vS f^{(k)} = [\vS f^{(k)}_1,\ldots, \vS f^{(k)}_{d_k}]$ is applied to each coordinate, and similarly for the non-linearities $\rho_k \circ f^{(k)}= [\rho_k \circ f^{(k)}_1, \ldots, \rho_k \circ f^{(k)}_{d_k}]$.

Then $Z^{(k)}$ is latent-convergent towards $f^{(k)}$, with a multiplicative constant that depends on the amplitude of the parameters. The rate is $\frac{1}{\sqrt{\alpha_n n}}$, both for random edges or weighted edges. By Prop. \ref{prop:transferconv}, transferability error follows with the same rate, and multiplicative constant that depends on the maximal amplitude of parameters. The detailed proof can be found in App. \ref{app:SGNN}, see Prop.~\ref{prop:Sconvadj}.

\subsubsection{Estimation error}

For the estimation error, we obtain for Assumption~\ref{ass:stability_estim_error} and \ref{ass:lossbounded}: for any $M>0$, there exist $C_M$ such that $p_n = n^{-M}$ and 
\begin{align*}
  \Delta^a_n &=\order{1/(\alpha_n n)},\quad 
  \Delta^A_n =\order{1},\quad 
  \Delta^z_n =\order{1},\quad 
  c_n =\order{1},\quad 
  C_n =\order{(1/\alpha_n)^L}
\end{align*}
Multiplicative constants depends polynomially on the amplitude of the parameters. Moreover, Lipschitzness for $d_{\oYy,n}$ can easily be derived (see App. \ref{app:SGNN}). As a result, the estimation error is indeed in $\order{R_\Theta\frac{\sqrt{\log n}}{\min(n \alpha_n, \sqrt{n})}}$ for a certain constant $R_\Theta$ that depends on the maximum amplitude and dimension of the parameters. The detailed proof can be found in App. \ref{app:SGNN}, see Prop.~\ref{prop:Sconvadj}.

\subsubsection{Universality with Positional Encodings}\label{app:univ}

Here we consider only the case $a_{ij} = w(x_i,x_j)$, and input Positional Encodings instead (we ignore eventual node features)
$$
z^{(0)}_i = \frac{1}{n} \sum_j f_{\theta'}(A_{i,j})
$$
where $f_{\theta'}:\RR\to \RR$ is an MLP. Note that $Z^{(0)}(A)$ is indeed permutation-equivariant wrt $A$.

This results in the same rate for transferability and estimation error as above, with $\alpha_n=1$, see App.~\ref{app:SGNN} for proof. The limit cGNN is as above, but with input function corresponding to the PEs:
$$
f^{(0)}(x) = \int f_{\theta'}(w(x,x')) dP_x(x')\, .
$$

Then, in \cite{Keriven2021} it is shown that we obtain universality in the following cases (see original paper for details):
\begin{itemize}
  \item \textbf{(Most) SBMs:} here $\Xx = \{1,\ldots, K\}$. If the probability matrix $C = [w(k,\ell)]_{k,\ell=1}^K$ is invertible, and the (finite) distribution $P_x$ is such that for all $s \in \{-1,0,1\}^K$, $s^\top P_x = 0$ implies $s=0$, then $\fF$ is universal for all functions. 
  \item \textbf{Additive kernel:} here $w(x,x')=v(u(x)+u(x'))$. If $u,v$ are continuous and \emph{injective} functions on their respective domains, then $\fF$ is universal for continuous functions. 
%
  \item \textbf{Unidimensional radial kernel}: here $\Xx = [-1,1]$ and $w(x,x') = v(|x-x'|)$. Assume $v$ is continuous and injective. If $P_x$ is symmetric (that is, $P_x([a,b]) = P_x([-b, -a])$ for all intervals), then $\fF$ is universal among continuous \textbf{and symmetric} functions.
  If $P_x$ is not symmetric, then $\fF$ is universal for all continuous functions. 
  \item \textbf{Spherical kernels}: If $\Xx = \mathbb{S}^{d-1}$ is the $d$-dimensional sphere and $w(x,x') = v(x^\top x')$ with continuous injective $v$, and $P_x$ has a density $p$ w.r.t. the uniform distribution on the sphere such that: the unique decomposition $p(x) = \sum_{k\geq 0} \sum_{j=1}^{N(d,k)} a_{k,j} Y_{k,j}(x)$ where $Y_{k,j}$ are spherical harmonics, is such that $x \mapsto [\sum_{j=1}^{N(d,k)} a_{k,j} Y_{k,j}(x)]_k$ is injective (see \cite{Keriven2021} and references therein for details on harmonic decomposition), 
  then $\fF$ is universal for all continuous functions. 
\end{itemize}

\subsection{GCN (Normalized Laplacian)}\label{app:ex_lapl}

Here the GNN has the same shape as above but with $S = D^{-1/2}A D^{-1/2}$ where $D = D(A) = \text{diag}(A1_n)$ is the diagonal matrix of degrees. Vanilla GCN \cite{Kipf2017} would add self-loop, this does not change the results. 

Here we consider the same LPM as above, but assume additionally that $d(x) = \int w(x,x') dP_x(x') \geq c_{\min} >0$ for all $x$. As before, $\alpha_n \gtrsim (\log n)/n$ but the multiplicative constant depends on $c_{\min}$, see \cite{Keriven2020a}.

\paragraph{Transferability}
Here the continuous-GNN is the same as above, but with operator 
$$
\vS f(x) = \int \frac{w(x,x')}{\sqrt{d(x)d(x')}} f(x') dP_x(x')
$$
The latent-convergence rate and transferability error are also the same, but the multiplicative constant also depends polynomially on $c_{\min}$. The detailed proof can be found in App. \ref{app:SGNN}, see Prop.~\ref{prop:Sconvlapl}.

\paragraph{Estimation error}
Similarily, the estimation error is also the same as in the previous example, but the multiplicative constant also depends polynomially on $c_{\min}$ and $p_n=0$ however. The detailed proof can be found in App. \ref{app:SGNN}, see Prop.~\ref{prop:Sconvlapl}.

\subsection{GAT}\label{app:ex_gat}

For the next three examples, we consider $a_{ij} = w(x_i,x_j)$ with Lipschitz $w$.

Graph Attention Network (GAT) were introduced by \cite{Velickovic2018}. Inspired by the attention mechanism in Transformers, \emph{attention coefficients} are computed along each edge of the graph at each layer, using the node representations. Then normalized sum-aggregation is performed with this coefficients:
\begin{align}
  \label{eq:gat}
  z_i^{(k)} &= \frac{\sum_{j=1}^n c^{(k)}\pa{z_i^{(k-1)}, z_j^{(k-1)}, a_{ij}} m^{(k)}(z_j^{(k-1)})}{\sum_{j=1}^n c^{(k)}\pa{z_i^{(k-1)}, z_j^{(k-1)}, a_{ij}}}
\end{align}
where $m^{(k)}:\RR^{d_{k-1}} \to \RR^{d_k}$ is some mapping (often an MLP), and $c^{(k)}(z_i^{(k)}, z_j^{(k)}, a_{ij}) \in \RR_+$ are attention coefficients that depend on the node representations at each end of the considered edge, and on the edge weight $a_{ij}$. Often, these coefficients are computed only along existing edges, such that $c(\cdot, \cdot, 0) = 0$. A common choice is to choose exponential coefficients, such that the aggregation operation is a soft-max, traditionally found in Transformers.

Assuming the attention coefficient are lower bounded and Lipschitz, latent-convergence has been proven in \cite{Cordonnier2023} with rate $\sqrt{(\log n)/n}$, for an architecture that uses layers with adapted non-linear operator
$$
f^{(k)}(x) = \int \frac{c^{(k)}(f^{(k-1)}(x), f^{(k-1)}(y), w(x,y))}{\int c^{(k)}(f^{(k-1)}(x), f^{(k-1)}(z), w(x,z))dP_x(z)} m^{(k)}(y) dP_x(y)
$$
More details on the proof can be found in App. \ref{app:agg}, see Prop.~\ref{prop:gatconv}.

\subsection{Generalized mean aggregation}\label{app:ex_genmean}

Generalized means \cite{generalized_mean_gnn} are defined as
\begin{equation}\label{eq:genmean}
  z_i^{(k)} = g\pa{\frac{1}{|\Nn(i)|} \sum_{j \in \Nn(i)} g^{-1}\pa{a_{ij} m^{(k)}(z_j^{(k-1)})}}
\end{equation}
for some invertible function $g$ (possibly defined on a bounded domain), generally applied component-wise. In this case, following the popular formalism from \cite{kolmogorov_mean}, in order to legitimately be considered as a ``generalized mean'', some regularity conditions are required on $g$, typically, it must be a strictly increasing continuous function. For instance, taking $g(x) = e^x$ (assuming positivity of the inputs for invertibility) yields the geometric mean, while taking $g(x) = x^{1 / p}$ (again, assuming nonnegativity of the message if necessary) yields a power mean as employed in the moment-based aggregations from \cite{Corso2020} (up to centering, omitted here for simplicity).

In \cite{Cordonnier2023}, for weighted edges $a_{ij} = w(x_i,x_j)$, latent-convergence has been proven towards an architecture that uses the continuous aggregation
\begin{equation}\label{eq:genmeancont}
  f^{(k)}(x) = g\pa{\int g^{-1}\pa{w(x,y) m^{(k)}(f^{(k-1)}(y))}dP_x(y)}
\end{equation}
For this, we need to assume that the message functions $m^{(k)}$ are Lipschitz and bounded, and that $g$ is $\alpha_g$-Hölder and bounded. The latent-convergence rate is then $O((\log n/n)^{\alpha_g \alpha_F^L/2})$ for some coefficient $\alpha_F$. We refer to App. \ref{app:agg} and Prop.~\ref{prop:genmeanconv} for details, but for instance:
\begin{enumerate}
  \item If $g(x) = e^x$ (geometric mean) on a bounded domain, $\alpha_g=\alpha_F = 1$.
  \item If $g(x) = x^{1/p}$ (moment-based aggregation), then $\alpha_g = 1/p$ and $\alpha_F=1$
  \item If $g^{-1}$ is Lipschitz (and $g$ is $\alpha_g$-Hölder), then $\alpha_F=\alpha_g$. 
\end{enumerate}

\subsection{Maximum aggregation}\label{app:ex_max}

The so-called maximum aggregation is mentioned in the literature \cite{Hamilton2017} but less common in practice. It is defined by
\begin{equation}\label{eq:max}
  z_i^{(k)} = \max_j\pa{a_{ij} m^{(k)}(z_j^{(k-1)})}
\end{equation}
where the max is taken componentwise along the dimension $d_k$.

In \cite{Cordonnier2023}, for weighted edges $a_{ij} = w(x_i,x_j)$, latent-convergence is proven with a rate $((\log n)/n)^{1/d_x}$ towards an architecture with continuous aggregation
\begin{equation}\label{eq:maxcont}
  f^{(k)}(x) = \sup_y\pa{w(x,y) m^{(k)}(f^{(k-1)}(y))}
\end{equation}
For this the measurable set $(\Xx, P_x)$ needs to satisfy a so-called \emph{volume retaining property}, see \cite{Cordonnier2023} for details.

\subsection{Invariant Graph Network (IGN)}

These networks use \emph{high-order equivariant operators} at each layer, which have been entirely characterized in the seminal paper of \cite{Maron2019}. On graphons, with proper normalization, latent-convergence has been proven by \cite{convergence_ign}, directly on weighted edges or through an additional estimator on random edges, which we skip here for simplicity. We refer to the original paper for architectural details.

The authors in \cite{convergence_ign} actually expresses a convergence in probability, such that the probability of exceeding $\epsilon$ is in $\frac{1}{\sqrt{n}\epsilon}$ (see \citep[Lemma 2]{convergence_ign}), but since all functions are bounded here this can be turned into a convergence in expectation\footnote{that is, $\EE err \lesssim \PP(err \geq \epsilon) + \epsilon \leq \frac{1}{\sqrt{n} \epsilon} + \epsilon$, then choose $\epsilon\sim \frac{1}{n^{1/4}}$} with rate $\frac{1}{n^{1/4}}$.

Note that the first convergence result in \cite{convergence_ign} for general IGN is in $L^2$ norm for graphons, which is slightly different from our definition of latent-convergence. The latter is more akin to their second result in MSE (which is more related to ``order-2'' latent-convergence which implies regular latent-convergence, see App.~\ref{app:general}), which is however only valid for a restricted class of IGN. We leave details for future work.

\subsection{Manifold NN}\label{app:ex_mnn}

Manifold NN were introduced by \cite{Wang2022b} as an equivalent to Convolutional NNs on manifold. The convergence result that we report here are from \cite{Wang_manifold_generalization}.

Here we assume that $\Mm \subset \Xx$ is a $d_\Mm$-dimensional manifold, and that $P_x$ is the uniform distribution on $\Mm$. The graph is built with the weighted edges 
$$
a_{ij} = \frac{1}{n t_n^{d/2+1} (4\pi)^{d/2}} e^{-\frac{-\norm{x_i-x_j}^2}{t_n}}
$$
and an MNN uses filters of the normalized Laplacian $L = \frac{1}{n t_n}(D-A)$ at each layer. Then, under regularity conditions on the filters and the input function $f^{(0)}$ that is sampled to obtain node features, $\pP$ convergence is shown with rate $\frac{\sqrt{\log n}}{n^{2/(d_\Mm+6)}}$. See \cite{Chew2022a} for details on hypotheses, assumptions, etc.

\section{Bonus: general approaches for GNNs}\label{app:general}

In this section, we provide general approaches for latent-convergence and (sometimes) estimation error for GNNs, based on existing results \cite{Keriven2023, Cordonnier2023}. In App.~\ref{app:SGNN} we start with GNNs that use an operator $S$ for aggregation of the form \eqref{eq:SGNN}, and provide latent-convergence and estimation error based on properties of the operator itself. Then in App.~\ref{app:agg}, we look at more general aggregation function, and provide a proof of latent-convergence based on regularity conditions developed by \cite{Cordonnier2023}. We leave the estimation error of such models for future work.

In what follows, we may need notions of latent-convergence with different metrics, mostly for technical convenience.
\begin{definition}[Latent-convergence of order $q$]\label{def:Pconvq}
  Let $h \in \Hh_{all}$ and $q\geq 1$. If there exists $f \in \Ff_{all}$ such that
  \begin{equation}
    \EE_{P_n} \pa{\frac{1}{n} \sum_{i=1}^n d_\oYy(h(A,Z)_i, f(x_i))^q}^{1/q} \leq r_n \xrightarrow[n \to \infty]{} 0
  \end{equation}
  we say that $h$ is latent-convergent \textbf{of order $q$} towards $f$ with rate $r_n$.
\end{definition}
By Prop.~\ref{prop:equivariance}, regular latent-convergence (Def.~\ref{def:Pconv}) is indeed latent-convergence of order $1$. A simple application of Hölder's inequality shows that for any $q\geq 1$, latent-convergence of order $q$ implies latent-convergence of order $1$. Note that, unlike order-1 latent-convergence, the expression here cannot be simplified by exchangeability, because of the power $(\cdot)^{1/q}$ inside the expectation. 

\paragraph{Notations.} The set $\Bb(\Xx)$ is the space of bounded function, equipped with the infinity norm. Multivariate functions $\Bb(\Xx)^d$ are equipped with $\norm{f}_\infty = \sup_x \norm{f(x)}_2$. For a collection $X=(x_1,\ldots, x_n)$ and $f\in \Bb(\Xx)^d$, we denote by $\iota_X f = [f(x_1),\ldots f(x_n)] \in\RR^{n \times d}$ the \emph{sampling operator}. For a vector $v$, we define $\norm{v}_q = (\sum_i |v_i|^q)^{1/q}$. For a square matrix $S$, $\norm{S}_{q} = \sup_v \norm{Sv}_q/\norm{v}_q$. In both cases $\norm{\cdot} = \norm{\cdot}_2$ for short. For a matrix $Z\in \RR^{n \times d}$, we also define the norm $\normMSEq{Z} = (\frac{1}{n} \sum_i \norm{Z_{:,i}}_2^q)^{1/q}$ and $\normMSEinfty{Z} = \sup_i \norm{Z_{i,:}}_2$.

\subsection{$S$-GNNs: aggregation with graph operator}\label{app:SGNN}

We first start by GNNs that use a graph operator \eqref{eq:SGNN} for aggregation, which we call $S$-GNNs in the rest. We recall that they are written 
\begin{equation}
  \label{eq:SGNN2}
  Z^{(k)} = \rho_k\pa{Z^{(k-1)}\theta_0^{(k)} + S Z^{(k-1)}\theta_1^{(k)} + 1_n {b^{(k)}}^\top} \in \RR^{n \times d_k}, \quad \Phi_\theta(A,Z) = Z^{(L)}
\end{equation}
where $S=S(A)$ is an equivariant graph operator (i.e. $S(\sigma A \sigma^top) = \sigma S(A) \sigma^top$) with weights $\theta^{(k)}_p$, bias $b^{(k)}$, and non-linearities $\rho_k$. In the whole section, the non-linearity will satisfy the following.
\begin{assump}\label{ass:sigma}
  The functions $\rho_k$ are $1$-Lipschitz, and $\abs{\rho_k(x)}\leq \abs{x}$.
\end{assump}

In this section, the input $Z^{(0)}$ are PEs (potentially only node features):
\begin{equation}\label{eq:PE2}
  Z^{(0)} = Z^{(0)}(A,Z)
\end{equation}
Here we first study latent-convergence for $S$-GNNs \eqref{eq:SGNN2}. It will entirely rely upon some notion of convergence \textbf{of the operator $S$ itself}, which is a well-known type of results for graph operators. We then move on to estimation error by bounded differences. Our approach is based on \cite{Keriven2023}.

\subsubsection{Latent-convergence and transferability error of $S$-GNNs}\label{sec:PconvSGNN}

As detailed above, in many cases, the matrix $S$ will converge to an \emph{operator} $\vS: \Bb(\Xx) \to \Bb(\Xx)$, where $\Bb(\Xx)$ is the space of bounded function on $\Xx$. 
When similarly the PEs $Z^{(0)}$ is latent-convergent to some $f^{(0)} \in \Bb(\Xx)^{d_0}$,
we can define the family of limit continuous GNNs (cGNN). 
As detailed in the previous Appendix, for a $S$-GNN $\Phi_\theta$ defined by \eqref{eq:SGNN2}, the \emph{same} set of parameters $\theta$ define a $\vS$-cGNN as 
\begin{equation}\label{eq:cSGNN}
  f^{(k)} = \rho_k\circ\pa{{\theta_0^{(k)}}^\top f^{(k-1)} + {\theta_1^{(k)}}^\top \vS f^{(k-1)} + b^{(k)}}, \quad \cPhi_\theta(\vS, f^{(0)}) = f^{(L)}
\end{equation}
where $f^{(k)}\in \Bb(\Xx)^{d_k}$ are multivariate functions, $\vS f^{(k)} = [\vS f^{(k)}_1,\ldots, \vS f^{(k)}_{d_k}]$ and $\rho_k \circ f^{(k)}= [\rho_k \circ f^{(k)}_1, \ldots, \rho_k \circ f^{(k)}_{d_k}]$.

Let us now state our assumptions on the convergence of individual elements. In the rest of the section, we consider some $q\geq 1$ to express latent-convergence of order $q$ (Def.~\ref{def:Pconvq}), which brings a measure of flexibility to the various assumptions that we need to prove in each example. While $q=1$ is sufficient, for $S$-GNNs $q=2$ will often be more convenient, as it involves the spectral norm. The convergence of the operator $S$ will be the following.

\begin{assump}[Latent-convergence of $S$]\label{ass:Sconv}
  For any $f \in \Bb(\Xx)$, assume that the function $(A,X) \mapsto S \iota_X f$ is latent-convergent of order $q$ towards $\vS f$ with rate $\norm{f}_\infty r_n^{(S,q)}$, where $r_n^{(S,q)} \to 0$.
\end{assump}
Remark that $(A,X) \mapsto S \iota_X f$ is a valid equivariant function that uses directly the (unobserved) latent variables $X$: it is mostly an abstract object that serves to express the convergence of $S$, and not something that is computed in practice.

Naturally, we also need convergence of the input node features/PE.
\begin{assump}\label{ass:PEconv}
  The Positional Encodings $Z^{(0)}(A,Z)$ are latent-convergent of order $q$ towards $f^{(0)} \in \Bb(\Xx)^{d_0}$ with rate $r^{(Z,q)}_n\to 0$.
\end{assump}
As mentioned in the main paper, the prototypical example of convergent input node features/PE is when the node features $Z=\iota_X f^{(0)}$ are themselves a sampling of some function over the latent variables, in which case $r^{(Z)}_n=0$. This is the choice we make in most examples in the main paper (except Sec.~\ref{sec:univ}).

Finally, various elements need to be bounded, which for some examples will only happen with high probability, similarly to Assumption \ref{ass:stability_estim_error}. The assumption below is generally mild.
\begin{assump}\label{ass:SZbound}
  For the same set of graphs $\Gg_n \subset \gG_n$ as Assumption \ref{ass:stability_estim_error}, assume that:
  \begin{equation}
    \normq{S} \leq\begin{cases}
      c_n^{(S,q)} &\text{if $(A,Z) \in \Gg_n$}\\ 
      C_n^{(S,q)} &\text{otherwise}
    \end{cases}, \quad 
    \normMSEq{Z^{(0)}} \leq\begin{cases}
      c_n^{(Z,q)} &\text{if $(A,Z) \in \Gg_n$}\\ 
      C_n^{(Z,q)} &\text{otherwise}
    \end{cases}
  \end{equation}
\end{assump}

Then we have the following result, presented here in a simplified version. A proof and detailed version is available in Appendix \ref{app:convmatrixMPdetail}.
\begin{theorem}[$\pP$-Convergence of $S$-GNN]\label{thm:convmatrixMP}
  Suppose that Assumptions \ref{ass:Sconv}, \ref{ass:PEconv} and \ref{ass:SZbound} hold. Denote by $p_n = P_{n}(\Gg_n^c)$ the probability that $(A,Z)\notin \Gg_n$. Then $\Phi_{\theta}$ is latent-convergent of order $q$ towards $\cPhi_\theta(\vS, f^{(0)})$, with rate 
  \begin{equation}\label{eq:transferrorMGNN}
    r^{(\theta)}_n \lesssim r^{(Z,q)}_n + r^{(S,q)}_n + p_n(1+C^{(Z,q)}_n)(1+ C^{(S,q)}_n)^L
  \end{equation}
  where the multiplicative constant depends polynomially on $\norm{\theta_i^{(k)}}$, $\norm{b^{(k)}}$, $\norm{f^{(0)}}_\infty$ and $c_n^{(S,q)}$.
\end{theorem}

As seen in the last Chapter, this means that such $S$-GNNs are indeed in $\Hhtrans$, and that under Assumption \ref{ass:loss} on the loss, their limit risk is the regular ML risk applied to the cGNN:
\begin{equation}
  \RrGMLn(\Phi_\theta) \xrightarrow[n \to \infty]{} \RrGML (\Phi_\theta) = \RrML(\cPhi_\theta(\vS, f^{(0)}))
\end{equation}
Moreover, by applying Prop.~\ref{prop:transferconv}, if we consider a set of $S$-GNNs $\Hh_\Theta = \{\Phi_\theta~|~ \theta \in \Theta\}$ with bounded parameters $\theta$, the \textbf{transferability error} is directly bounded:
\begin{equation}\label{eq:SGNNtransfererror}
  \sup_{\theta \in \Theta} \abs{\RrGMLn(\Phi_\theta) - \RrGML(\Phi_\theta)} \lesssim \sup_\theta r^{(\theta)}_n \to 0
\end{equation}
where the multiplicative constant is either the Lipschitz constant of the loss, or depends on $\Theta$ for the square loss (see Prop.~\ref{prop:transferconv} and Lemma \ref{lem:boundsGNN} in Appendix). 


\subsubsection{Estimation error of $S$-GNNs}

In the main paper, we bound the estimation error with a stability bound (Assumption \ref{ass:stability_estim_error}) and a covering numbers approach (Thm.~\ref{thm:stabilitycovering}).
As before, Assumption \ref{ass:stability_estim_error} can be deduced from a stability property directly on $S$ and $Z^{(0)}$. 
We state this in the assumption below, which is (rather pointlessly, but unavoidably) verbose. 

\begin{assump}\label{ass:stabilitymatrixMP}
  In the following, we denote by $S=S(A)$, $S' = S(A')$, $Z^{(0)} = Z^{(0)}(A,Z)$ and ${Z'}^{(0)} = Z^{(0)}(A',Z')$. Assume that there is a set of graphs $\Gg_n$ such that, for all $(A,Z), (A', Z') \in \Gg_n$:
  \begin{itemize}
    \item if $Z=Z'$ and $A, A'$ differs on only one coordinate $A_{ij}$, 
    \begin{equation*}
      \sum_{\ell,k=1}^n \abs{S_{\ell k} - S'_{\ell k}} \leq \Delta^{a,S}_n, \quad \sum_k \norm{Z^{(0)}_{k,:} - {Z'}^{(0)}_{k,:}} \leq \Delta^{a,Z}_n
    \end{equation*}
    \item if $Z=Z'$, and $A,A'$ differs on one line $A_{i,:}$ and the corresponding column $A_{:,i}$,
    \begin{equation*}
      \sum_{\ell,k=1}^n \abs{S_{\ell k} - S'_{\ell k}} \leq \Delta^{A,S}_n, \quad \sum_k \norm{Z^{(0)}_{k,:} - {Z'}^{(0)}_{k,:}} \leq \Delta^{A,Z}_n
    \end{equation*}
    \item if $A=A'$ and $Z,Z'$ differs on one sample $z_i \neq z'_i$,
    \begin{equation*}
      \sum_k \norm{Z^{(0)}_{k,:} - {Z'}^{(0)}_{k,:}} \leq \Delta^{z,Z}_n
    \end{equation*}
  \end{itemize}  
  
\end{assump}

The proof of the following result, and a more detailed version, is in the Appendix.

\begin{theorem}\label{thm:SGNNstab}
  Let $\Phi_\theta$ be a $S$-GNN. Suppose that Assumption \ref{ass:stabilitymatrixMP} holds, and that Assumption \ref{ass:SZbound} holds for $q=1$ and $q=\infty$. Suppose that the loss satisfies Assumption \ref{ass:loss}. Then Assumption \ref{ass:stability_estim_error} and \ref{ass:lossbounded} are satisfied, with
  \begin{align*}
    \Delta^{a}_n &\lesssim  \Delta^{a,S}_n + \Delta^{a,Z}_n \\
    \Delta^{A}_n &\lesssim   \Delta^{A,S}_n + \Delta^{A,Z}_n\\
    \Delta^{z}_n &\lesssim  \Delta^{z,Z}_n \\
    c_n^{(\ell)} &= \order{1} \\
    C_n^{(\ell)} &\lesssim (1+C_n^{(S,\infty)})^L (1+ C_n^{(Z,\infty)})
  \end{align*}
  where the multiplicative constants depend polynomially on $\norm{\theta^{(k)}_0}, \norm{\theta^{(k)}_1}, \norm{b^{(k)}}$, $c_n^{(S,1)}, c_n^{(S,\infty)}$, and $c_n^{(Z,\infty)}$. 
\end{theorem}

Now let $\Hh_\Theta = \{ \Phi_\theta ~|~ \theta \in \Theta\}$ be a set of $S$-GNNs with parameters in a bounded set $\Theta$. To apply theorem \ref{thm:stabilitycovering} and bound the estimation error, we need to bound its covering numbers. For this we introduce the metric between parameters $\theta, \theta'$:
\begin{equation}
  \norm{\theta}_\Theta = \max_k \max\pa{\norm{\theta_0^{(k)}}, \norm{\theta_1^{(k)}}, \norm{b^{(k)}}}
\end{equation}
and we suppose that $\max_{\theta \in \Theta} \norm{\theta}_\Theta \leq D_\Theta$. Then we have the following.

\begin{proposition}\label{prop:sgnn_covering}
  Suppose that Assumption \ref{ass:stabilitymatrixMP} holds, and that Assumption \ref{ass:SZbound} holds for $q=1$ and $q=\infty$. Then
  \begin{equation*}
    \Nn(\Hh_\Theta, \epsilon, d_{\oYy,n}) \lesssim \pa{\frac{D}{\epsilon}}^{dim(\theta)}
  \end{equation*}
  where $dim(\theta)$ is the number of parameters in the GNN, and $D$ is polynomial is in $O(D_\Theta^{2L} {C_n^{(S,\infty)}}^{2L} C_n^{(Z,\infty)})$.
\end{proposition}

Hence, applying Theorem \ref{thm:stabilitycovering}, assuming that $p_n$ is negligible, we obtain that the rate for the estimation error of $S$-GNNs is approximately
\begin{equation}\label{eq:estimerror_sgnn_rate}
  \pa{\Delta^{A,S}_n + \Delta^{A,Z}_n + \sqrt{n}(\Delta^{a,S}_n + \Delta^{a,Z}_n)}  \sqrt{\frac{\dim(\theta)\log n}{n}}
\end{equation}
With a multiplicative constant that depends polynomially on the amplitude of the parameters and constants $c^{(\cdot)}_n$. We apply it to our examples in the next section.

\subsubsection{Examples of $S$-GNNs}
In this section we apply the results above to our two examples of $S$-GNNs, and obtain detailed versions of their results in App.~\ref{app:gnn}. For simplicity we omit the PEs.

Recall that to bound the estimation \emph{and} tranferability error, we need to satisfy:
\begin{itemize}
  \item Assumptions \ref{ass:Sconv} and \ref{ass:SZbound} for the same $q$. Here, we will consider $q=2$. This will yield latent-convergence of order $q$, and recall that order $2$ implies order $1$.
  \item Assumptions \ref{ass:SZbound} for $q=1$ and $q=\infty$ (technically, for $q=1$ only the constant $c^{(S,1)}$ is required), and Assumption \ref{ass:stabilitymatrixMP}, which bounds the estimation error.
\end{itemize}


\begin{proposition}[Transferability and Estimation error for Normalized Adjacency]\label{prop:Sconvadj}
  Consider LPM random graphs with iid latent positions $x_i \sim P_x$ and Bernoulli edges $a_{ij} \sim \text{Ber}(\alpha_n w(x_i,x_j))$, with $\alpha_n \gtrsim \frac{\log n}{n}$.
  Take $S = A/(\alpha_n n)$, and define 
  \begin{equation}\label{eq:continuousadj}
    \vS f(x) = \int w(x,x') f(y) dP_x(x')\, .
  \end{equation}
  Then
  \begin{itemize}
    \item Assumption \ref{ass:Sconv} is satisfied with $q=2$ and
    $
    r_n^{(S,2)}=\frac{1}{\sqrt{\alpha_n n}} + \sqrt{\frac{d_x}{n}}
    $
    \item For any $M$, there is a constant $C_M$ such that Assumption \ref{ass:SZbound} is satisfied for $S$ with $p_n = n^{-M}$ and $c_n^{(S,2)} = 2$, $C_n^{(S,2)}=C_M/\alpha_n$.
    \item For the same $p_n$ as above, Assumption \ref{ass:SZbound} is satisfied with $c^{(S,1)} = c^{(S,\infty)} = \order{1}$, and $C^{(S,\infty)} = 1/\alpha_n$.
    \item for the same $p_n$ as above, Assumption \ref{ass:stabilitymatrixMP} is satisfied with $\Delta^{a,S}_n = 1/(n \alpha_n)$ and $\Delta^{(A,S)}_n = \order{1}$.
  \end{itemize}
  We apply Thm.~\ref{thm:convmatrixMP} and \ref{thm:SGNNstab} to obtain the final rates written in App.~\ref{app:gnn} and Tab.~\ref{tab:example}.

  For weighted edges $a_{ij} = w(x_i,x_j)$, the results are the same but with $p_n=0$ and $\alpha_n = 1$.
\end{proposition}

The proof relies on concentration for matrices with Bernoulli entries \cite{Lei2015}, to get rid of the random edges, then convergence towards continuous operators, which can be shown e.g. by chaining \cite{Keriven2020a}. It is given in App.~\ref{app:Sconvadj} for completeness.

\begin{proposition}[Transferability and Estimation error for Laplacian]\label{prop:Sconvlapl}
  Consider LPM random graphs with iid latent positions $x_i \sim P_x$ and Bernoulli edges $a_{ij} \sim \text{Ber}(\alpha_n w(x_i,x_j))$. Assume that $d_w(x) = \int w(x,y)dP_x(y) \geq c_{\min} > 0$, and $\alpha_n \gtrsim \frac{\log n}{c_{\min}^2 n}$.
  
  Take $S = D^{-\frac12}AD^{-\frac12}$, and define
  \begin{equation}\label{eq:continuouslapl}
    \vS f(x) = \int_\Xx \frac{w(x,x')}{\sqrt{d_w(x) d_w(x')}} f(x') dP_x(x')\, .
  \end{equation}

  Then
  \begin{itemize}
    \item Assumption \ref{ass:Sconv} is satisfied with $q=2$ and
    $
    r_n^{(S,2)}=c_{\min}^{-2}\pa{\frac{1}{\sqrt{\alpha_n n}} + \sqrt{\frac{d_x}{n}}}
    $
    \item Assumption \ref{ass:SZbound} is satisfied for $S$ and $q=2$ with $c_n^{(S,2)} = 1$, $p_n = 0$.
    \item for any $M$, there is a constant $C_M$ such that Assumption \ref{ass:SZbound} is satisfied for $S$ with $p_n = n^{-M}$ and $c^{(S,1)} = c^{(S,\infty)} = C_M/c_{\min}$.
    \item for the same $p_n$ as above, Assumption \ref{ass:stabilitymatrixMP} is satisfied with $\Delta^{a,S}_n = O(1/(c_{\min} n \alpha_n))$ and $\Delta^{(A,S)}_n = O(1/c^2_{\min})$.
  \end{itemize} 

  As a consequence, the transferability error and estimation error are of the same order as Prop.~\ref{prop:Sconvadj} for the adjacency matrix, up to some multiplicative factor depending on $c_{\min}$.

  For weighted edges $a_{ij} = w(x_i,x_j)$, the results are the same but with $\alpha_n = 1$.
\end{proposition}

The proof is similar, using a concentration bound for Laplacian that we developed in \cite{Keriven2020}. It is given in App.~\ref{app:Sconvlapl} for completeness.

\subsubsection{Positional Encodings}

To complete the proof of Sec.~\ref{sec:univ}, we need to show latent-convergence and stability for the PE \eqref{eq:PE2} for weighted edges $a_{ij} = w(x_i,x_j)$. By easy proofs (see \cite{Keriven2021}), we obtain 
\begin{itemize}
    \item latent-convergence with $r_n^{(Z,2)} = \order{1/\sqrt{n}}$ towards
    $$
    f^{(0)}(x) = \int f_{\theta'}(w(x,x'))dP_x(x')
    $$
    and a multiplicative constant that depends on $\theta'$.
    \item bounds $c_n^{(Z,\infty)} \leq \sup_{t \in [0,1]}\norm{f_{\theta'}(t)} = \order{1}$, $C_n^{(Z,\infty)}$ does not matter since $p_n=0$,
    \item stability $\Delta^{(Z,a)}_n = \order{1/n}$, $\Delta_n^{(Z,A)} = \order{1}$, $\Delta_n^{(Z,z)}=0$ (since node features are ignored here).
\end{itemize}
We can conclude to find the results of App.~\ref{app:univ}.

\subsection{General Aggregation}\label{app:agg}

We now study general aggregation functions of the form 
\begin{equation}\label{eq:MPwithF}
  z_i^{(k)} = F^{(k-1)}(z_i^{(k-1)}, \{(z_j^{(k-1)}, a_{ij})\}_{i=1}^n), \quad \Phi(A,Z) = Z^{(L)}
\end{equation}
where $F^{(k)}$ are aggregation functions that treat the set of neighbor representations $\{(z_j^{(k-1)}, a_{ij})\}_{i=1}^n$ as a (multi)set, i.e., in a permutation-invariant manner.
We will show a general approach for latent-convergence, based on an approach from \cite{Cordonnier2023}. 

%
Crucially, the theory that we develop here is valid only for \emph{non-random weighted edges} $a_{ij} = w(x_i,x_j)$, and we assume $w(\cdot)\geq w_{\min}$ for some $w_{\min}$ that will vary depending on the example. We also consider the regular node features case $z_i^{(0)} = f^{(0)}(x_i)$. OUr notations largely follow \cite{Cordonnier2023}. 
%



\paragraph{Definition of the limit.} Unlike the previous $S$-GNN, for general MP the very definition of the ``limit'' cGNN is more convoluted than simply replacing the discrete operator $S$ by its continuous counterpart. 
We note that, \emph{if} the node representations at layer $k-1$ \emph{were} a sampling of a function $Z^{(k-1)}=\iota_X f$, then the layer $k$ would be of the form:
\begin{equation*}
  z_i^{(k)} = F^{(k-1)}(f(x_i), \{(f(x_j), w(x_i,x_j))\}_{j\neq i})
\end{equation*}
since $a_{ij} = w(x_i,x_j)$. Hence, when, $n \to \infty$, we consider limit functions of the form 
\begin{equation}\label{eq:vF}
  f^{(k)}(x) := \vF^{(k-1)}\pa{f^{(k-1)}(x), \pa{f^{(k-1)}, w(x, \cdot)}}, \quad \cPhi = f^{(L)}
\end{equation}
where $\pa{f^{(k)}, w(x, \cdot)}$ is a shortcut notation for the map $x' \mapsto \pa{f^{(k)}(x'), w(x, x')}$.

By stacking such layers, we can recursively build a ``continuous GNN'' which propagates functions over the latent space, akin to replacing $S$ with an operator in the $S$-GNN case.

We will proceed in two steps: \emph{define} the limit by assuming convergence of the expectation of $F$, then prove a large deviation bound with McDiarmid inequality, by assuming bounded difference with respect to each $x_i$. The next definition is a simplified version of Def. 5 from \cite{Cordonnier2023}.

\begin{definition}[latent-convergence of MP layers]
  Let $F$ be an GNN layer with input features of dimension $d_k$. We say that $F$ is latent-convergent towards $\vF$ with rate $r_n(\cdot)$ if, for all $f\in\Bb(x)^{d_k}$,
  \begin{equation}
    \sup_x\norm{\EE_{x_i}F\pa{f(x), \{f(x_i), w(x, x_i)\}_{i=2}^n} - \vF(f(x), (f, w(x,\cdot)))} \leq r_n(f) \to 0
  \end{equation}
\end{definition}
Note that here, this is not the ``vanilla'' definition of latent-convergence from the last chapter, but a specialized definition. The supremum over $x$, in particular, indicates that we are reasoning with the stronger (strongest) ``uniform'' convergence, rather that any order-$q$ convergence in expectation as before. Again, it may be possible to obtain different versions of these results, but we stick with those from \cite{Cordonnier2023} in this manuscript.

\paragraph{Convergence of GNN} Consider a sequence of layers $F^{(1)}, \ldots, F^{(L)}$, and assume that they are latent-convergent towards $\vF^{(k)}$ with rate $r_n^{(k)}(\cdot)$. Denote by $f^{(k)}$ the intermediate functions of the continuous GNN \eqref{eq:vF}. We are now ready to state our convergence theorem: under a bounded difference assumption and mild stability bounds, we will have latent-convergence of the GNN. The notion of bounded difference is a bit verbose, in the sense that we authorize some transform of the function before computing  the stability bounds. This is a very generic notion, but in practice we introduce this especially to deal with generalized mean, see \cite{Cordonnier2023}.

\begin{assump}[Generalized bounded difference]\label{ass:MPboundeddiff}
  There is a function $\psi$, assumed invertible (on the proper domain) such that: for all $k$,
  \begin{enumerate}
    \item For all $x_i=x'_i$ except for $i=2$, we have the bounded difference
    \begin{align}
      \sup_x \Big\Vert \psi^{-1}& \pa{F^{(k)}\pa{f^{(k-1)}(x), \{f^{(k-1)}(x_i), w(x,x_i)\}_{i\geq 2}}} \\
      &\qquad- \psi^{-1}\pa{F^{(k)}\pa{f^{(k-1)}(x), \{f^{(k-1)}(x'_i), w(x,x'_i)\}_{i\geq 2}}} \Big \Vert \leq \Delta^{F}_n
    \end{align}
    \item We also have bounded difference in expectation:
    \begin{align}
      \sup_x \Big\Vert \psi& \pa{ \EE_{x_i} \psi^{-1} \pa{F^{(k)}\pa{f^{(k-1)}(x), \{f^{(k-1)}(x_i), w(x,x_i)\}_{i\geq 2}}}} \\
      &\qquad- \EE_{x_i}\pa{F^{(k)}\pa{f^{(k-1)}(x), \{f^{(k-1)}(x'_i), w(x,x'_i)\}_{i\geq 2}}} \Big \Vert \leq \tilde \Delta^{F}_n
    \end{align}
    \item The function $\psi$ is $\alpha_\psi$-Hölder on its domain with $0<\alpha_\psi\leq 1$:
    \begin{equation}
      \norm{\psi(y)-\psi(y')}\leq K_\psi \norm{y-y'}^{\alpha_\psi}
    \end{equation}
    \item Finally, we also need a global Lipschitz bound: there exists an exponent $0<\alpha_F\leq 1$ such that, for all $z_i, z'_j$, and $w_{\min}\leq w_i\leq 1$:
    \begin{equation}
      \norm{F^{(k)}(z_i, \{z_j, w_j\}_{j=2}^n) - F^{(k)}(z'_i, \{z'_j, w_j\}_{j=2}^n)} \lesssim \max_i \norm{z_i-z'_i}^{\alpha_F}
    \end{equation}
  \end{enumerate}
\end{assump}



Then, the theorem is the following.

\begin{theorem}[{\citep[Thm. 15]{Cordonnier2023}}, simplified]\label{thm:GNNconv}
  Consider a sequence of layers $F^{(1)}, \ldots, F^{(L)}$, and assume that they are latent-convergent towards $\vF^{(k)}$ with rate $r_n^{(k)}$. Denote $r_n = \max_k r^{(k)}(f^{(k-1)})$. Under Assumptions \ref{ass:MPboundeddiff}, the GNN $\Phi$ is latent-convergent of order $\infty$ towards $\cPhi$ with rate 
  \begin{equation}\label{eq:GNNconv}
    r_n^{(\Phi)} \lesssim \left[ \pa{ (\Delta^F_n)^2 n \log n}^{\alpha_\psi/2} + \tilde \Delta^F_n + r_n\right]^{\alpha_F^{L-1}}
  \end{equation}
  where the multiplicative constant depends polynomially on all the quantities in the problem.
\end{theorem}
The version in \citep[Thm. 15]{Cordonnier2023} is a high-probability result rather than convergence in expectation, with a dependency on the probability of failure $\delta$ that is in $\log(1/\delta)$, so it is easy to obtain a result in expectation.

As before, an immediate corollary is that a set of GNNs $\Hh$ with bounded parameters converges to the set of the corresponding cGNNs $\Ff_\Hh$ with uniform rate $\sup_\Phi r^{(\Phi)}_n$. Unlike the previous examples of $S$-GNN, where the rate in $\frac{1}{\sqrt{n \alpha_n}}$ depended on the sparsity, with the classical $1/\sqrt{n}$ for dense graphs, here the convergence rate depends on multiple factors, that must be studied on a case-by-case basis: \emph{a minima}, we need $\tilde \Delta^F_n \to 0$ and $\Delta^F_n = o(\sqrt{n \log n})$ to obtain a rate that goes to $0$. This is the case for all of our examples except for max aggregation, which requires a dedicated proof, see \cite{Cordonnier2023}. 

\paragraph{Examples} We now apply the approach above to GAT and generalized mean. The details can be found in \cite{Cordonnier2023}.

\begin{proposition}[Convergence for GAT]\label{prop:gatconv}
  Consider the GAT example. Define the continuous counterpart of \eqref{eq:gat}:
  \begin{equation*}
    f^{(k)}(x) = \int \frac{c^{(k)}(f^{(k-1)}(x), f^{(k-1)}(y), w(x,y))}{\int c^{(k)}(f^{(k-1)}(x), f^{(k-1)}(z), w(x,z))dP_x(z)} m^{(k)}(y) dP_x(y)
  \end{equation*}
  Assume that the message functions $m^{(k)}$ are Lipschitz and bounded. Assume that the attention coefficients are lower bounded $c(x,y,t)\geq c_{\min} >0$, and are Lipschitz $\abs{c(x,y,t) - c(x',y',t)} \lesssim \norm{x-x'} + \norm{y-y'}$. Then the hypotheses of Theorem \ref{thm:GNNconv} are satisfied with $\Delta^F_n=O(1/n),~\tilde  \Delta^F_n = 0,~r_n=O(1/\sqrt{n})$, and $\alpha_\psi=\alpha_F=1$. The final rate is therefore $O(\sqrt{\log n / n})$.
\end{proposition}

\begin{proposition}[Convergence for generalized mean] \label{prop:genmeanconv}
  Consider the generalized mean example. Define the continuous counterpart of \eqref{eq:genmean}:
  \begin{equation*}
    f^{(k)}(x) = g\pa{\int g^{-1}\pa{w(x,y) m^{(k)}(f^{(k-1)}(y))}dP_x(y)}
  \end{equation*}
  Assume that the message functions $m^{(k)}$ are Lipschitz and bounded. Assume that $g$ is $\alpha_g$-Hölder and bounded, and that Assumption \ref{ass:MPboundeddiff} is satisfied for some $\alpha_F$ (see examples below). Then the hypothese of Theorem \ref{thm:GNNconv} are satisfied with $\Delta^F_n=O(1/n),~\tilde \Delta^F_n = O(1/n^{\alpha_g/2}),~r_n=O(1/n^{\alpha_g/2})$, and $\alpha_\psi=\alpha_g$. The final rate is therefore $O((\log n/n)^{\alpha_g \alpha_F^L/2})$.

  For $\alpha_g, \alpha_F$, we give several examples:
  \begin{enumerate}
    \item If $g(x) = x^{1/p}$ (moment-based aggregation), then $\alpha_g = 1/p$ and $\alpha_F=1$
    \item If $g^{-1}$ is Lipschitz (and $g$ is $\alpha_g$-Hölder), then $\alpha_F=\alpha_g$. Note that for geometric mean on a bounded domain, $g$ is also Lipschitz, such that $\alpha_g=1$.
  \end{enumerate}
\end{proposition}

\paragraph{Other examples.}
Max aggregation \cite{Cordonnier2023}, IGN \cite{convergence_ign}, and Manifold NNs \cite{Wang_manifold_generalization} are not covered by this general theory, we refer the reader to the original papers for details.

\section{Proofs of Appendix \ref{app:general}}

In this last section we gather the technical proofs related to App.~\ref{app:general}, and other additional material.

\subsection{Proof and detailed version of Theorem \ref{thm:convmatrixMP}}\label{app:convmatrixMPdetail}

\begin{theorem}[Theorem \ref{thm:convmatrixMP}, detailed]\label{thm:convmatrixMPdetailed}
  Suppose that Assumptions \ref{ass:Sconv}, \ref{ass:PEconv} and \ref{ass:SZbound} hold. Denote by $p_n = P_{n}(\Gg_n^c)$ the probability that $(A,Z)\notin \Gg_n$. Then $\Phi_{\theta}$ is latent-convergent of order $q$ towards $\Phi_\theta(\vS, f^{(0)})$, with rate 
  $$
  r_n^{(\theta)} \lesssim \sum_{\ell=0}^{L-1} \left[ \prod_{p=\ell+1}^{L-1} A_p\right] B_\ell + r^{(Z,q)}_n\prod_{\ell=0}^{L-1} A_\ell
  $$
  where
  \begin{align*}
    A_k &= \norm{\theta_0^{(k)}} + \norm{\theta_1^{(k)}} c_n^{(S,q)} \\
    B_k &= \norm{\theta_1^{(k)}}\pa{p_n C_n^{(S,q)}(D_k + D'_k) + r^{(S,q)}_n D'_k} \\
    D_k &= \sum_{\ell=1}^{k} \pa{\prod_{\ell=l}^{k-1} u_p}\norm{b^{(\ell)}} + \pa{\prod_{p=0}^{k-1} u_p} C^{(Z,q)}_n \\
    D'_k &= \sum_{\ell=1}^{k} \pa{\prod_{\ell=l}^{k-1} u'_p}\norm{b^{(\ell)}} + \pa{\prod_{p=0}^{k-1} u'_p} \norm{f^{(0)}}_\infty \\
    u_k &= \norm{\theta_0^{(k)}} + C^{(S,q)}_n\norm{\theta_1^{(k)}}, \quad u'_k = \norm{\theta_0^{(k)}} + \norm{\vS}_{\infty}\norm{\theta_1^{(k)}}
  \end{align*}
  where $\norm{\vS}_\infty = \sup_{f} \norm{\vS f}_\infty/\norm{f}_\infty$.
\end{theorem}

\begin{proof}
  We bound recursively $E_k = \EE_{P_n} \normMSEq{Z^{(k)} - \iota_X f^{(k)}}$. Using the Lipschitz property of $\rho_k$ and Lemma \ref{lem:boundsq},
  \begin{align*}
    E_k &= \EE_{P_n} \normMSEq{Z^{(k)} - \iota_X f^{(k)}} \\
    & = \EE_{P_n} \Big\Vert\rho_k\pa{Z^{(k-1)}\theta_0^{(k-1)} + S Z^{(k-1)}\theta_1^{(k-1)} + 1_n {b^{(k)}}^\top} \\
    &\qquad\qquad - \rho_k\pa{(\iota_X f^{(k-1)}) \theta_0^{(k-1)} + (\iota_X \vS f^{(k-1)})\theta_1^{(k-1)} + 1_n {b^{(k)}}^\top} \Big\Vert_{q,2,n} \\
    & \leq \norm{\theta_0^{(k-1)}} E_{k-1} + \norm{\theta_1^{(k-1)}}\EE_{P_n}\normMSEq{S Z^{(k-1)} - (\iota_X \vS f^{(k-1)})}
  \end{align*}
  From there, again with Lemma \ref{lem:boundsq}
  \begin{align*}
    \EE_{P_n}\normMSEq{S Z^{(k-1)} - (\iota_X \vS f^{(k-1)})} &\leq \EE_{P_n}\normq{S}\normMSEq{Z^{(k-1)} - \iota_X f^{(k-1)}} \\
    &\qquad + \underbrace{\EE_{P_n}\normMSEq{S \iota_X f^{(k-1)} - (\iota_X \vS f^{(k-1)})}}_{\leq r_n^{(S,q)} \norm{f^{(k-1)}}_\infty}
  \end{align*}
  Hence, the first term is bounded by decomposing under the event $\Ee_n = (\normq{S}\leq c^{(S,q)})$, by Assumption \ref{ass:SZbound}:
  \begin{align*}
    \EE_{P_n}\normq{S}\normMSEq{Z^{(k-1)} - \iota_X f^{(k-1)}} &\leq c^{(S,q)}\EE_{P_n}\pa{\normMSEq{Z^{(k-1)} - \iota_X f^{(k-1)}} ~\Big|~ \Ee_n} \\
    &\qquad+ p_n C^{(S,q)}_n (D_{k-1} + \norm{f^{(k-1)}}_\infty)
  \end{align*}
  where $D_k$ is the maximum value of $\normMSEq{Z^{(k)}}$ under $\Ee_n^c$.

  Finally, we have naturally
  $$
  \EE_{P_n}\pa{\normMSEq{Z^{(k-1)} - \iota_X f^{(k-1)}} ~\Big|~ \Ee_n} \leq \EE_{P_n}\pa{\normMSEq{Z^{(k-1)} - \iota_X f^{(k-1)}}} = E_{k-1}
  $$
  which gives us the desired recursion. We conclude using Lemma \ref{lem:boundsGNN} to bound $D_k$ and $\norm{f^{(k)}}_\infty$.
\end{proof}

\subsection{Proof and detailed version of Theorem \ref{thm:SGNNstab}}\label{app:SGNNstabdetail}

\begin{theorem}[Detailed version of Thm.~\ref{thm:SGNNstab}]\label{thm:SGNNstabdetailed}
  Let $\Phi_\theta$ be a $S$-GNN. Suppose that Assumption \ref{ass:stabilitymatrixMP} holds. Then Assumption \ref{ass:stability_estim_error} is satisfied, with
  \begin{align}
    \Delta^{a}_n &\lesssim  \pa{\sum_{\ell=1}^k \left[ \prod_{p=\ell}^{k-1} u^1_k \right] C_k} \Delta^{a,S}_n + \pa{\prod_{\ell=1}^{k-1} u^1_\ell} \Delta^{a,Z}_n \\
    \Delta^{A}_n &\lesssim  \pa{\sum_{\ell=1}^k \left[ \prod_{p=\ell}^{k-1} u^1_k \right] C_k} \Delta^{A,S}_n + \pa{\prod_{\ell=1}^{k-1} u^1_\ell} \Delta^{A,Z}_n\\
    \Delta^{z}_n &\lesssim  \pa{\prod_{\ell=1}^{k-1} u^1_\ell} \Delta^{z,Z}_n
  \end{align}
  where
  \begin{align*}
    u^q_k &= \norm{\theta^{(k)}_0} + c^{(S,q)}_n \norm{\theta^{(k)}_1} \\
    C_k &= \sum_{\ell=1}^k \left[ \prod_{p=\ell}^{k-1} u^\infty_k \right] \norm{b^{(k)}} + \pa{\prod_{\ell=1}^{k-1} u^\infty_\ell} c_n^{(Z,\infty)}
  \end{align*}

  For simplicity we do not give the details for Assumption \ref{ass:lossbounded} here.
\end{theorem}

\begin{proof}
  Let $(A,Z)$, $(A',Z') \in \Gg_n$ be two graphs, $S,S'$ their associated operator, and $Z^{(k)}, {Z'}^{(k)}$ the intermediate layers propagating by the GNN. Denote by $E_k = \sum_i \norm{Z^{(k)}_i - {Z'}^{(k)}_i}$. We compute
  \begin{align*}
    E_k &\leq \norm{\theta^{(k)}_0} \sum_i \norm{Z^{(k-1)}_i - {Z'}^{(k-1)}_i} \\
    &\quad+ \norm{\theta_1^{(k)}} \pa{\normMSEinfty{Z^{(k)}}\sum_{ij}\abs{S_{ij} - S'_{ij}} + \norm{S'}_1 \sum_i \norm{Z^{(k-1)}_i - {Z'}^{(k-1)}_i}} \\
    &= u^{1}_k E_{k-1} + C_k \abs{S_{ij} - S'_{ij}}
  \end{align*}
  where $C_k = \normMSEinfty{Z^{(k)}}$, which by Lemma \ref{lem:boundsGNN} is bounded using $\norm{S}_\infty \leq c_n^{(S,\infty)}$ since $(A,Z) \in \Gg_n$. A simple recursion concludes the proof.

  Assumption \ref{ass:lossbounded} is shown in the same manner, we do not detail the proof here.
\end{proof}

\subsection{Proof and detailed version of Prop \ref{prop:sgnn_covering}}

\begin{proof}[Proof of \ref{prop:sgnn_covering}]
  Our strategy is to show that $d_{\oYy,n}(\Phi_\theta, \Phi_{\theta'}) \leq D_n \norm{\theta - \theta'}_\Theta$, such that $\Nn(\Hh, \epsilon, d_{\oYy, n}) \leq \Nn(\Theta, \epsilon/D_n, \norm{\cdot}_\Theta) \lesssim (D_\Theta D_n/\epsilon)^{dim(\Theta)}$, where the last inequality is the classical covering numbers of a compact set of finite dimension.

  For any $A,Z \in \gG_n$, denote by $Z^{(k)}$ the layer of the GNN with parameters $\theta$ and ${Z'}^{(k)}$ the layers with parameters $\theta'$, and $E_k = \normMSEinfty{Z^{(k)} - {Z'}^{(k)}}$. Note that the PEs are the same $Z^{(0)} = {Z'}^{(0)}$. We have, using Lemma \ref{lem:boundsq},
  \begin{align*}
    E_k &\leq \max_i \norm{\rho_k\pa{\theta^{(k)}_0 Z^{(k)}_{i,:} + \theta^{(k)}_1 \sum_j S_{ij} Z^{(k)}_{j,:} + b^{(k)}} - \rho_k\pa{{\theta'}^{(k)}_0 {Z'}^{(k)}_{i,:} + {\theta'}^{(k)}_1 \sum_j S_{ij} {Z'}^{(k)}_{j,:} + {b'}^{(k)}}} \\
    &\leq \normMSEinfty{Z^{(k)}} \norm{\theta_0^{(k)} - {\theta'}_0^{(k)}} + \norm{{\theta'}^{(k)}_0} E_{k-1} \\
    &\quad + \norm{S}_\infty \pa{\normMSEinfty{Z^{(k)}} \norm{\theta_1^{(k)} - {\theta'}_1^{(k)}} + \norm{{\theta'}^{(k)}_1} E_{k-1}} + \norm{b^{(k)} - {b'}^{(k)}} \\
    &\leq u_\infty E_{k-1} + C_k \norm{\theta - \theta'}_\Theta
  \end{align*}
  where $u_\infty = D_\Theta(1+\norm{S}_\infty)$ and using Lemma \ref{lem:boundsGNN}, $C_k = (D_\Theta + C_n^{(Z,\infty)}) u_\infty^k$. We obtain a global bound in $D_\Theta^{L} \norm{S}_\infty^{L} C_n^{(Z,\infty)}$, which concludes the proof.
\end{proof}

\subsection{Proof and detailed version of Prop \ref{prop:Sconvadj}}\label{app:Sconvadj}

\begin{proof}[Proof of Prop.~\ref{prop:Sconvadj}]
  Let $W = [w(x_i,x_j)]$. By \cite{Lei2015}, for any $M$ there is a constant $C_M$ such that with probability $1- n^{-M}$,
  \begin{equation*}
    \norm{S - W/n} \leq C_M/\sqrt{n \alpha_n}
  \end{equation*}
  and thus, since $\norm{S - W/n} \lesssim 1/\alpha_n$ with probability $1$,
  \begin{equation*}
    \EE_{P_n} \norm{S - W/n} \lesssim \frac{n^{-M}}{\alpha_n} + \frac{C_M}{\sqrt{n \alpha_n}} \lesssim \frac{1}{\sqrt{n \alpha_n}}
  \end{equation*}
  Hence 
  $$
  \normMSE{(S-W/n)\iota_X f} \leq \norm{S-W/n}\normMSE{\iota_X f} \lesssim \frac{\norm{f}_\infty}{\sqrt{n \alpha_n}}
  $$

  Moreover, by \citep[Lem. 4]{Keriven2020a}, for any $f \in \Bb(\Xx)$, with probability at least $1-\delta$
  \begin{equation*}
    \normMSE{W\iota_X f - \iota_X \vS f} \lesssim \frac{\sqrt{d_x} + \sqrt{\log(1/\delta)}}{\sqrt{n}}
  \end{equation*}
  which similarly yields
  $$
  \EE_{P_n}\normMSE{W\iota_X f - \iota_X \vS f} \lesssim \sqrt{\frac{d_x}{n}}
  $$
  which concludes the proof. Assumption \ref{ass:SZbound} is similarly shown with \cite{Lei2015}.

  Then, we have $\norm{S}_1 = \norm{S_\infty} = \sup_i \sum_j \abs{S_{ij}} = \max_i d_i/(n \alpha_n)$ (since $S$ is symmetric), where $d_i$ are the degrees. Under the same probability as above, by \citep[Lemma 3]{Keriven2020} the degrees are all such that
  \begin{equation}\label{eq:degreesLei}
    \forall i,\quad c_{\min}n \alpha_n \leq d_i \leq n \alpha_n
  \end{equation}
  where we recall that $c_{\min}$ lower-bounds the degre \emph{function}. Hence with probability $1-p_n$, $c^{(S,1)} = O(1)$.
  
  Finally, Assumption \ref{ass:stabilitymatrixMP} is direct, again by using bounds on the degrees \eqref{eq:degreesLei} for $\Delta^{A}_n$.
\end{proof}

\subsection{Proof and detailed version of Prop \ref{prop:Sconvlapl}}\label{app:Sconvlapl}

\begin{proof}[Proof of Prop.~\ref{prop:Sconvlapl}]
  The proof is similar to that of Prop.~\ref{prop:Sconvadj}, but with \citep[Thm.~4]{Keriven2020} to bound $\norm{S - D_K^{-1/2} K D_K^{1/2}}$ where $D_K$ is the diagonal degrees of $K$, and \citep[Lem. 5]{Keriven2020a} to bound the second term in the triangular inequality.

  The rest of the proof is similar, bounding the degrees with \eqref{eq:degreesLei}, and using for assumption \ref{ass:stabilitymatrixMP}
  \begin{align}
    \sum_{ij} \abs{S_{ij} - S'_{ij}} &= \sum_{ij} \abs{\frac{A_{ij}}{\sqrt{d_id_j}} - \frac{A'_{ij}}{\sqrt{d'_i d'_j}}}\notag \\
    &= \sum_{ij} \abs{\frac{A_{ij} - A'_{ij}}{\sqrt{d_id_j}}} + \sum_{ij} \abs{A'_{ij}\pa{\frac{1}{\sqrt{d_id_j}} - \frac{1}{\sqrt{d'_i d'_j}}}}\notag \\
    &= \sum_{ij} \abs{\frac{A_{ij} - A'_{ij}}{\sqrt{d_id_j}}} + \sum_{ij} \abs{A'_{ij}\pa{\frac{\sqrt{d_i d_j} - \sqrt{d'_i d'_j}}{\sqrt{d_i d_j d'_i d'_j}}}}\label{eq:laplconv_intermediate}
  \end{align}
  The first term is easy to bound using $d_i = O(n \alpha_n)$.

  For the second term, when $A_{ij} = A'_{ij}$ except for $i_0,j_0$, we have $d_i=d'_i$ except for $i=i_0$ or $j_0$ for which $d_i - d'_i = 1$, such that
  \begin{align*}
    \sum_{ij} \abs{A'_{ij}\pa{\frac{ \sqrt{d_i d_j} - \sqrt{d'_i d'_j}}{\sqrt{d_i d_j d'_i d'_j}}}} &= \sum_{i} \abs{A'_{ij_0}\pa{\frac{\sqrt{d_i}(\sqrt{d_{j_0}} - \sqrt{d'_{j_0}})}{\sqrt{d_i^2 d_{j_0} d'_{j_0}}}}} \\
    &\quad + \sum_{j} \abs{A'_{i_0j}\pa{\frac{\sqrt{d_j}(\sqrt{d_{i_0}} - \sqrt{d'_{i_0}})}{\sqrt{d_j^2 d_{i_0} d'_{i_0}}}}}
  \end{align*}
  and
  \begin{equation*}
    \abs{\sqrt{d_i}(\sqrt{d_{j_0}} - \sqrt{d'_{j_0}})} = \frac12 \sqrt{\frac{d_i}{d}}
  \end{equation*}
  for some $d = O(n \alpha_n)$, such that $\sqrt{\frac{d_i}{d}} = O(1)$. We conclude with $\sum_i A_{i j_0} = O(n \alpha_n)$ and $\sqrt{d_j^2 d_{i_0} d'_{i_0}} = O(n^2 \alpha_n^2)$.

  When $A$ and $A'$ differ by one line and one column $i_0$, the reasoning is the same, except that $d_i - d'_i = 1$ when $i \neq i_0$, and $d_{i_0} - d'_{i_0} = O(n \alpha_n)$ otherwise.

  The first term in \eqref{eq:laplconv_intermediate} is again easily bounded by bounding the degrees. The second term involves
  $$
  \sum_{j} \abs{A'_{i_0 j}\pa{\frac{\sqrt{d_{i_0} d_j} - \sqrt{d'_{i_0} d'_j}}{\sqrt{d_i d_j d'_id'_j}}}}
  $$
  which is easily in $O(1)$ by bounding all the degrees and using $\sum_{j} A'_{i_0 j} = d'_{i_0}$. Similarly when $j=i_0$.

  The remaining term is more involved. Note that, when $i\neq i_0$, the event $d_i - d'_i = 1$ can only happen for $O(\alpha_n n)$ nodes, since this is the number of $1$'s in both $A_{i_0,:}$ and $A'_{i_0,:}$. Define $s_i = 1$ when this happens and $0$ otherwise, such that $\sum_i s_i = O(n \alpha_n)$. By the same reasoning as above, we have
  $$
  \sqrt{d_i d_j} - \sqrt{d'_i d'_j} = \sqrt{d_i}(\sqrt{d_j} - \sqrt{d'_j}) + \sqrt{d'_j}(\sqrt{d_i} - \sqrt{d'_i}) = O(s_i + s_j)
  $$
  Finally,
  $$
  \sum_{ij} A'_{ij} (s_i + s_j) = d'_i \sum_j s_j + d'_j \sum_i s_i = O(n^2 \alpha_n^2)
  $$
  and since $\sqrt{d_i d_j d'_id'_j} = O(n^2\alpha_n^2)$ we have indeed $C^{(A,S)}=O(1)$.
\end{proof}

\subsection{Additional Technical Material}
\begin{lemma}\label{lem:boundsGNN}
  We have 
  \begin{equation}
    \normMSEq{Z^{(k)}} \leq \sum_{l=1}^{k} \pa{\prod_{p=l}^{k-1} u_p}v_l + \pa{\prod_{p=0}^{k-1} u_p} \normMSEq{Z^{(0)}}
  \end{equation}
  where $u_k = \norm{\theta_0^{(k)}} + \normq{S}\norm{\theta_1^{(k)}}$ and $v_k = \norm{b^{(k)}}$.
  
  
  Similarly,
  \begin{equation}
    \norm{f^{(k)}}_\infty \leq \sum_{l=1}^{k} \pa{\prod_{p=l}^{k-1} u_p}v_l + \pa{\prod_{p=0}^{k-1} u_p} \norm{f^{(0)}}_\infty
  \end{equation}
  where $u_k = \norm{\theta_0^{(k)}} + \norm{\vS}_{\infty \to \infty}\norm{\theta_1^{(k)}}$.
\end{lemma}
\begin{proof}
  Recursively, using $\abs{\sigma(t)} \leq \abs{t}$, Lemma \ref{lem:boundsq} and $\normMSEq{1_n b^\top} = \norm{b}$,
  \begin{align*}
    \normMSEq{Z^{(k)}} = \norm{\rho_k\pa{Z^{(k-1)}\theta_0^{(k-1)} + S Z^{(k-1)}\theta_1^{(k-1)} + 1_n {b^{(k)}}^\top}} \leq u_{k-1} \normMSEq{Z^{(k-1)}} + v_{k}
  \end{align*}
  Hence the result. 
  The proof for $f^{(k)}$ is similar.
\end{proof}

\begin{lemma}
  \label{lem:boundsq}
  For $X \in \RR^{n \times d}$, $\theta \in \RR^{d \times d'}$ and $S \in \RR^{n \times n}$ with $S_{ij}\geq 0$, we have
  \begin{align}
    \normMSEq{X\theta} &\leq \normMSEq{X}\norm{\theta} \\
    \normMSEq{SX} &\leq \normq{S}\normMSEq{X}
  \end{align}
\end{lemma}
\begin{proof}
  The first inequality is obvious:
  \begin{equation*}
    \normMSEq{X\theta} = \pa{\frac{1}{n}\sum_{i=1}^n \norm{\theta^\top X_{i,:}}^q}^{1/q} \leq \norm{\theta}\pa{\frac{1}{n}\sum_{i=1}^n \norm{X_{i,:}}^q}^{1/q} = \normMSEq{X}\norm{\theta}
  \end{equation*}
  For the second inequality, we denote by $v = [\norm{X_{i,:}}] \in \RR_+^n$, since $S_{ij}\geq 0$ we have
  \begin{align*}
    \normMSEq{SX} &= \pa{\frac{1}{n}\sum_{i=1}^n \norm{\sum_j S_{ij} X_{j,:}}^q}^{1/q} \\
    &\leq \pa{\frac{1}{n}\sum_{i=1}^n \pa{\sum_j S_{ij} \norm{X_{j,:}}}^q}^{1/q} = \frac{1}{n^{1/q}}\normq{Sv} \leq \frac{1}{n^{1/q}}\normq{S}\normq{v} = \normq{S}\normMSEq{X}
  \end{align*}
\end{proof}

\begin{theorem}
  [Generalized McDiarmid's inequality \cite{Combes2024}] \label{thm:mcdiarmid}
  Let $(X_1,\ldots, X_n) \in \Xx = \prod_{i=1}^n \Xx_i$ be independent random variables, and $f:\Xx \to \RR$. Assume that there is $\Yy \in \Xx$ such that: for all $X,X' \in \Yy$ such that $X_j = X'_j$ except for some index $i$,
  $$
  \abs{f(X)-f(X')} \leq c_i 
  $$
  and denote $p = \PP(X\in \Yy)$. Moreover, assume that $\abs{f(X)} \leq C$ with probability $1$. Then with probability at least $1-\delta - p$, 
  \begin{equation}
    \abs{f(X) - \EE f(X)} \lesssim p \pa{\norm{c}_1 + C} + \norm{c}_2 \sqrt{\log(1/\delta)}
  \end{equation}
  where $c=(c_1,\ldots, c_n)$.
\end{theorem}






\end{document}